\def\year{2020}\relax
\documentclass[letterpaper]{article} 
\usepackage{aaai20}  
\usepackage{times}  
\usepackage{helvet} 
\usepackage{courier}  
\usepackage[hyphens]{url}  
\usepackage{graphicx} 
\usepackage{graphicx}  
\usepackage{amsmath}
\usepackage{amssymb}
\usepackage{multirow}
\usepackage{svg}
\usepackage{makecell}
\usepackage{enumitem}
\usepackage{tabularx}
\usepackage{fixltx2e} 
\usepackage{booktabs} 
\usepackage{xcolor}
\usepackage{scalerel}
\usepackage{url}
\usepackage[toc, page]{appendix}

\def\ConvColor{rgb:yellow,5;red,2.5;white,5}
\def\ConvColorHeadC{rgb:blue,3;green,3}
\def\ConvColorHeadR{rgb:blue,3;green,1.5}
\def\ConvColorHeadCtr{rgb:red,3}
\def\XcorrColor{rgb:blue,5;red,5}

\title{SiamFC++: Towards Robust and Accurate Visual Tracking \\ with Target Estimation Guidelines}

\author{
Yinda Xu,\textsuperscript{\rm 1}\thanks{These authors contributed equally to this work.}\thanks{This work has been done when these authors were doing an intership at Megvii Inc.}
Zeyu Wang,\textsuperscript{\rm 2}\footnotemark[1]\footnotemark[2]
Zuoxin Li,\textsuperscript{\rm 2}
Ye Yuan,\textsuperscript{\rm 2}
Gang Yu\textsuperscript{\rm 2}\thanks{Corresponding Author}
\\
\textsuperscript{\rm 1}College of Electrical Engineering, Zhejiang University\\ 
\textsuperscript{\rm 2}Megvii Inc.\\
yinda\_xu@zju.edu.cn, wangzeyu0408@outlook.com, \{lizuoxin, yuanye, yugang\}@megvii.com 
}

\begin{document}
\maketitle

\begin{abstract}
Visual tracking problem demands to efficiently perform robust classification and accurate target state estimation over a given target at the same time. Former methods have proposed various ways of target state estimation, yet few of them took the particularity of the visual tracking problem itself into consideration. Based on a careful analysis, we propose a set of practical guidelines of target state estimation for high-performance generic object tracker design. Following these guidelines, we design our Fully Convolutional Siamese tracker++ (SiamFC++) by introducing both classification and target state estimation branch (\textbf{G1}), classification score without ambiguity (\textbf{G2}), tracking without prior knowledge (\textbf{G3}), and estimation quality score (\textbf{G4}). Extensive analysis and ablation studies demonstrate the effectiveness of our proposed guidelines. Without bells and whistles, our SiamFC++ tracker achieves state-of-the-art performance on five challenging benchmarks(OTB2015, VOT2018, LaSOT, GOT-10k, TrackingNet), which proves both the tracking and generalization ability of the tracker. Particularly, on the large-scale TrackingNet dataset, SiamFC++ achieves a previously unseen AUC score of 75.4 while running at over 90 FPS, which is far above the real-time requirement. Code and models are available at: \url{https://github.com/MegviiDetection/video_analyst}.
\end{abstract}

\section{Introduction}

\begin{figure}[t] 
    \centering
    \includegraphics[width=.95\columnwidth]{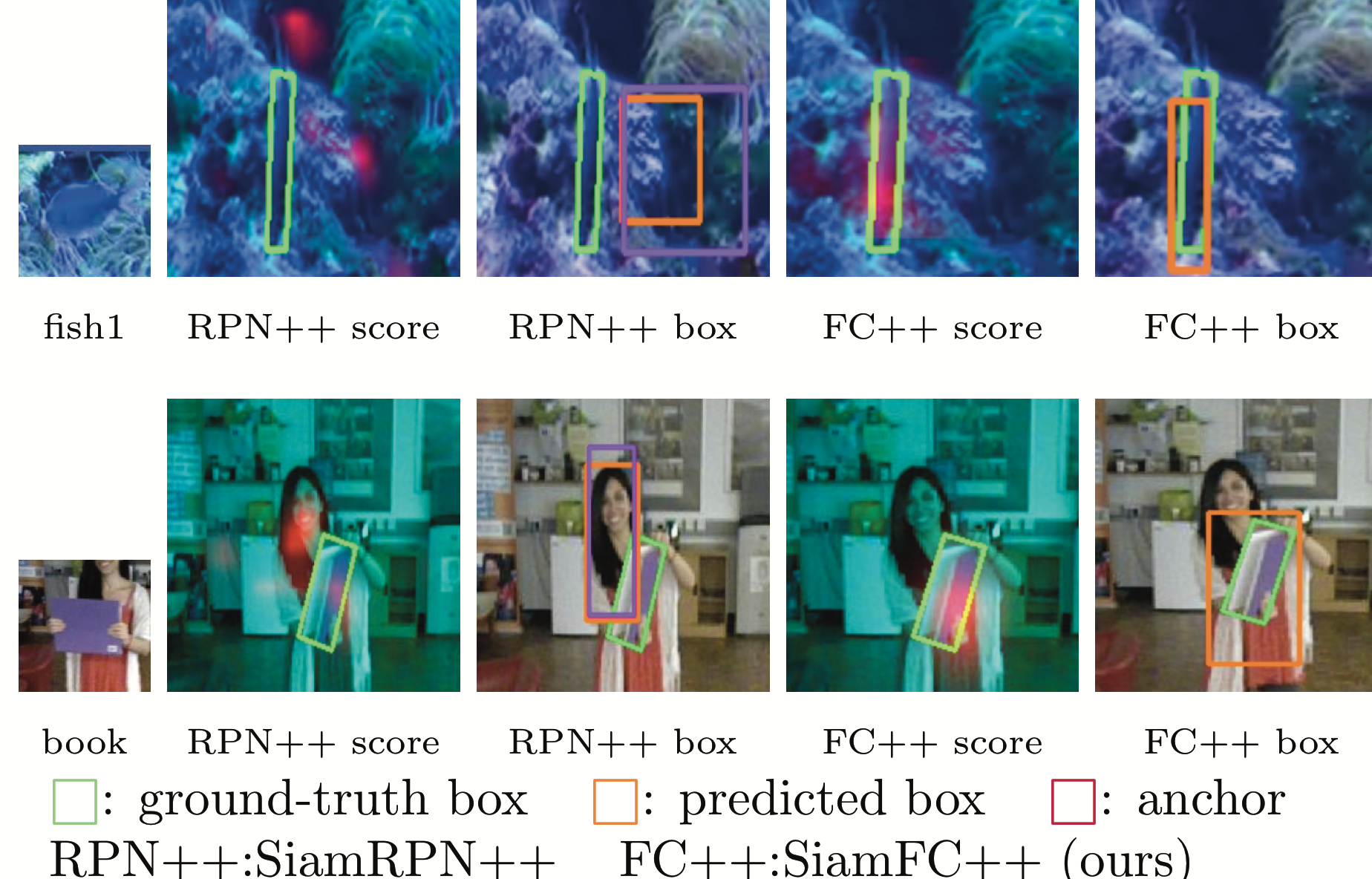}
    \caption{A comparison of our approach (following the guidelines) with state-of-the-art SiamRPN++ tracker (which violates some of the guidelines). Score maps are visualized by red color (i.e. red parts represent regions with high scores and vice versa). In the case of a significant change of target appearance, SiamRPN++ fails due to anchor-object mismatch while our SiamFC++ successes by directly matching between objects. See Section \ref{subsec:anchor-based analysis} for analysis in detail.}
    \label{fig:comparison anchor-based/anchor-free}
\end{figure}

Generic Visual Tracking aims at locating a moving object sequentially in a video, given very limited information, often only the annotation of the first frame. Being a fundamental build block in various areas of computer vision, the task comes with a variety of applications such as UAV-based monitoring~\cite{mueller2016benchmark} and surveillance system~\cite{kokkeby2015methods}. One unique characteristic of generic object tracking is that no prior knowledge (e.g., the object class)
about the object, as well as its surrounding environment, is allowed~\cite{huang2018got}.

Tracking problem can be treated as the combination of a \textit{classification} task and an \textit{estimation} task~\cite{danelljan2019atom}. The first task aims at providing a robust coarse location of the target via classification. The second task is then to estimate an accurate target state, often represented by a bounding box.
While modern trackers have achieved significant progress, surprisingly their methods for the second task (i.e. target state estimation) largely differ. Based on this aspect, previous methods can be roughly divided into three categories. The first category, including Discriminative Correlation Filter (DCF)~\cite{henriques2014high-speed,bolme2010visual} and SiamFC~\cite{bertinetto2016fully}, employs brutal multi-scale test which is inaccurate~\cite{danelljan2019atom} and inefficiency~\cite{li2018high}. Also, the prior assumption that target scale/ratio changes in a fixed rate in adjacent frames often does not hold in reality. For the second category, ATOM~\cite{danelljan2019atom} iteratively refines multiple initial bounding boxes via gradient ascending to estimate the target bounding box~\cite{jiang2018acquisition}, which yields a significant improvement on accuracy. However, this target estimation method brings not only a heavy computation burden but also many additional hyper-parameters (e.g. the number of initial boxes, distribution of initial boxes) that requires careful tuning. The third category is SiamRPN tracker family~\cite{li2018high,zhu2018distractor,li2019siamrpn++} that performs an accurate and efficient target state estimation by introducing the Region Proposal Network (RPN)~\cite{ren2015faster}. However, the pre-defined anchor settings not only introduce ambiguous similarity scoring that severely hinders the robustness (see Section \ref{subsec:anchor-based analysis}) but also need access to prior information of data distribution, which is clearly against the spirit of generic object tracking~\cite{huang2018got}.

Motivated by the aforementioned analysis, we propose a set of guidelines for high-performance generic object tracker design: 
\begin{itemize}[itemsep=0mm]
    \item \textbf{G1: decomposition of classification and state estimation} The tracker should perform two sub-tasks: classification and state estimation. Without a powerful classifier, the tracker cannot discriminate the target from background or distractors, which severely hinders its robustness~\cite{zhu2018distractor}. Without an accurate estimation result, the accuracy of the tracker is fundamentally limited~\cite{danelljan2019atom}.
    Those brutal multi-scale test approaches largely ignore the latter task, suffering from inefficiency and low accuracy.

    \item \textbf{G2: non-ambiguous scoring} The classification score should represent the confidence score of target existence directly, in the "field of view", i.e. sub-window of the corresponding pixel, rather than the pre-defined settings like anchor boxes. As a negative example, matching between objects and anchors (e.g. the anchor-based RPN branch) is prone to deliver a false positive result, leading to tracking failure (see Section \ref{subsec:anchor-based analysis} for more details).
    
    \item \textbf{G3: prior knowledge-free} Tracking approaches should be free of prior knowledge like scale/ratio distribution, as is proposed by the spirit of generic object tracking~\cite{huang2018got}. Dependency on prior knowledge of data distribution exists widely in existing methods, which hinders the generalization ability.
    
    \item \textbf{G4: estimation quality assessment} As is shown in previous researches~\cite{jiang2018acquisition,tian2019fcos}, using classification confidence for bounding box selection directly will result in degenerated performance. An estimation quality score independent of classification should be used, as in many previous pieces of research about both object detection and tracking~\cite{jiang2018acquisition,tian2019fcos,danelljan2019atom}. The astonishing accuracy of the second branch (e.g. ATOM and DiMP) largely comes from this guideline. While the others still overlook it, leaving room for further estimation accuracy improvement. 
\end{itemize}

Following the guidelines above, we design our SiamFC++ method based on fully-convolutional siamese trackers~\cite{bertinetto2016fully}, where each pixel of the feature map directly corresponds to each translated sub-window on the search image due to its fully convolutional nature. We add a regression head for accurate target estimation, in parallel with the classification head (\textbf{G1}). Since the pre-defined anchor settings is removed, the matching ambiguity (\textbf{G2}) and prior knowledge (\textbf{G3}) about target scale/ratio distribution is also removed. Finally, following \textbf{G4}, an estimation quality assessment branch is added to privilege bounding boxes with high quality.

Our contribution can be summarized in three-fold:
\begin{enumerate}[itemsep=0mm]
\item By identifying the unique characteristics of tracking, we devise a set of practical guidelines of target state estimation for modern tracker design.
\item We design a simple but powerful SiamFC++ tracker with the application of our proposed guidelines. Extensive experiments and comprehensive analyses demonstrate the effectiveness of our proposed guidelines.
\item Our approach achieves state-of-the-art results on five challenging benchmarks. To the best of our knowledge, our SiamFC++ is the first tracker that achieves an AUC score of 75.4 on the large-scale TrackingNet dataset~\cite{muller2018trackingnet} while running at over 90 FPS. 
\end{enumerate}

\begin{figure*}[t]
  \centering
  \includegraphics[page=1,width=.84\linewidth]{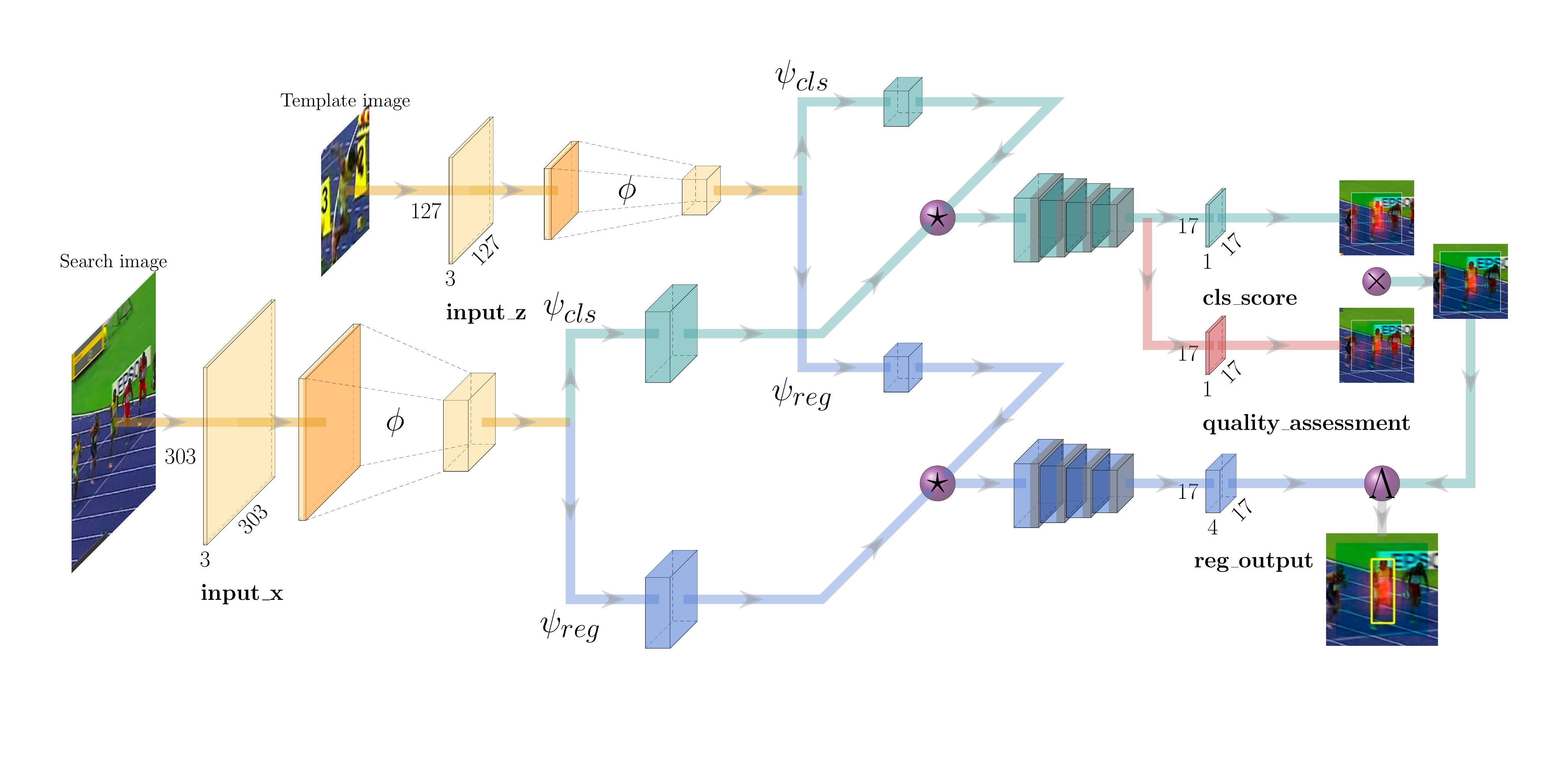}
  \begin{tabular}{r@{: }l r@{: }l r@{: }l r@{: }l}
  \textcolor{\ConvColor}{$\blacksquare$} & feature extractor & 
  \textcolor{\ConvColorHeadC}{$\blacksquare$} & classification branch & 
  \textcolor{\ConvColorHeadR}{$\blacksquare$} & regession branch &
  \textcolor{\ConvColorHeadCtr}{$\blacksquare$} & quality assessment \\
  \textcolor{\XcorrColor}{$\bullet$} & operation &
  $\star$ & cross-correlation &
  $\times$ & element-wise production &
  $\Lambda$ & argmax (taking \textit{left} w.r.t. \textit{right}) \\
  \end{tabular}
  \caption{Our SiamFC++ pipeline (AlexNet version). Boxes denote feature maps\iffalse while right branded boxes denote feature maps with ReLU activation\fi. The intermediate layers in the common feature extractor have been omitted for clarity. For score visualization, deep green color represents the corresponding region on input image of score map, while the brightness of red color denotes the magnitude of scores (min-max normalized). Better viewed in color with zoom-in.}
  \label{fig:SiamFC++ pipeline}
\end{figure*}

\section{Related Works}

\subsection{Tracking Framework}
\par
Modern trackers can be roughly divided into three branches by their way of target state estimation.
 
Some of them, including DCF~\cite{henriques2014high-speed,bolme2010visual} and SiamFC~\cite{bertinetto2016fully}, use multi-scale test to estimate the target scale. Concretely, by rescaling the search patch into multiple scales and assembling a mini-batch of scaled images, the algorithm picks the scale corresponding to the highest classification score as the predicted target scale in the current frame.  This strategy is fundamentally limited since bounding box estimation is inherently a challenging task, requiring a high-level understanding of the pose of objects~\cite{danelljan2019atom}.

Inspired by DCF and IoU-Net~\cite{jiang2018acquisition}, ATOM~\cite{danelljan2019atom}tracks target by sequential classification and estimation. The coarse initial location of the target obtained by classification is iteratively refined for accurate box estimation. The Multiple random initializations of bounding boxes in each frame and multiple back propagations in iterative refinement greatly slows down the speed of ATOM. This approach yields a significant improvement on accuracy but also brings a heavy computation burden. What's more, ATOM introduces many additional hyper-parameters that require careful tuning.

Another branch, named SiamRPN and its succeeding works~\cite{li2018high,zhu2018distractor,li2019siamrpn++} append a Region Proposal Network after a siamese network, achieving a previously unseen accuracy. RPN regresses the location shift and size difference between pre-defined anchor boxes and target location. However, the RPN structure is much more fit for object detection, in which a high recall rate is required, while in visual tracking one and only one object should be tracked. Also, the ambiguous matching between anchor box and object severely hinders the robustness of tracker (see Section \ref{subsec:anchor-based analysis}). Finally, the anchor setting does not comply with the spirit of generic object tracking, requiring pre-defined hyper-parameters describing its shape.

\subsection{Detection Framework}
With many unique characteristics, visual tracking task still has a lot in common with object detection, which makes each one task benefiting from each other possible. For example, the RPN structure first devised in Faster-RCNN~\cite{ren2015faster} achieves astonishing accuracy in SiamRPN~\cite{li2018high}. Inheriting from Faster-RCNN~\cite{ren2015faster}, most state-of-the-art modern detectors, named anchor-based detectors, have adopted the RPN structure and the anchor boxes setting~\cite{ren2015faster,liu2016ssd,Li_2018_ECCV}. The anchor-based detectors classifies pre-defined proposals called anchor as positive or negative patches, with an extra offsets regression to refine the prediction of bounding box locations. However, hyper-parameters introduced by anchor boxes (e.g. the scale/ratio of anchor boxes) have shown a great impact on the final accuracy, and require heuristic tuning~\cite{cai2018cascade,tian2019fcos}. Researchers have tried various ways to design anchor-free detectors, like predicting bounding boxes at points near the center of objects~\cite{redmon2016you,huang2015densebox}, or detecting and grouping a pair of corners of a bounding box~\cite{law2018cornernet}. In this paper, we show that a simple pipeline based on a carefully designed guidelines for target state estimation inspired by \cite{huang2015densebox,yu2016unitbox,tian2019fcos} can achieve state-of-the-art tracking performance.

\section{SiamFC++: Fully Convolutional Siamese Tracker for Object Tracking}
\label{sec:SiamFC++}
In this section, we describe our Fully Convolutional Siamese tracker++ framework in detail. Our SiamFC++ is based on SiamFC and progressively refined according to the proposed guidelines. As shown in Figure \ref{fig:SiamFC++ pipeline}, the SiamFC++ framework consists of a siamese subnetwork for feature extraction and a region proposal subnetwork for both classification and regression. 

\subsection{Siamese-based Feature Extraction and Matching}
\label{Siamese-based feature extraction and matching}
Object tracking task can be viewed as a \textit{similarity learning} problem~\cite{li2018high}. Concretely speaking, a siamese network is trained offline and evaluated online to locate a \textit{template} image within a larger \textit{search} image. A siamese network consists of two branches. The \textit{template} branch takes target patch in the first frame as input (denoted as $z$), while the \textit{search} branch takes the current frame as input (denoted as $x$). The siamese backbone, which shares parameters between two branches, performs the same transform on the input $z$ and $x$ to embed them into a common feature space for subsequent tasks.  A cross-correlation between template patch and search patch is performed in the embedding space $\phi$: 
\begin{equation}
    f_i(z,x) = 
    \psi_i\left(\phi\left(z\right)\right) \star
    \psi_i\left(\phi\left(x\right)\right)
    , i\in \left\{ \text{cls}, \text{reg} \right\}
    \label{eq:siamese formula}
\end{equation}
where $\star$ denotes the cross-correlation operation, $\phi(.)$ denotes the siamese backbone for common feature extraction, $\psi_i(.)$ denotes the task-specific layer and $i$ denotes the sub-task type ("cls" for classification and "reg" for regression). In our implementation, We use two convolution layers for both $\psi_{\text{cls}}$ and $\psi_{\text{reg}}$ after common feature extraction to adjust the common features into task-specific feature space. Note that the extracted features of $\psi_{\text{cls}}$ and $\psi_{\text{reg}}$ are of the same size.

\subsection{Application of Design Guidelines in Head Network}
Based on SiamFC, we progressively refine each part of our trackers following our guidelines.

\label{subsec:cls/reg heads}
Following \textbf{G1}, we design both classification head and regression head after the cross-correlation in the embedding space. For each pixel in feature maps, the classification head takes $\psi_{\text{cls}}$ as input and classifies the corresponding image patch as either one positive or negative patch, while the regression head takes $\psi_{\text{reg}}$ as input and outputs an extra offsets regression to refine the prediction of bounding box locations. The structure of heads is presented after the cross-correlation operation of Figure \ref{fig:SiamFC++ pipeline}. 

Specifically, for classification, location $(x,y)$ on feature map $\psi_{\text{cls}}$ is considered as a positive sample if its corresponding location $\left(\left\lfloor\frac{s}{2}\right\rfloor+ x s,\left\lfloor\frac{s}{2}\right\rfloor+ y s\right)$ on the input image falls into the ground-truth bounding box. Otherwise, it is a negative sample. Here $s$ is the total stride of backbone ($s=8$ in this paper). For the regression target of each positive location $(x,y)$ on feature map $\psi_{\text{reg}}$, the final layer predicts the
distances from the corresponding location $\left(\left\lfloor\frac{s}{2}\right\rfloor+ x s,\left\lfloor\frac{s}{2}\right\rfloor+ y s\right)$ to the four sides of the ground-truth bounding box, denoted as a 4D vector $\boldsymbol{t}^{*}=\left(l^{*}, t^{*}, r^{*}, b^{*}\right)$. Hence, the regression targets for location $(x,y)$ can be formulated as
\begin{equation}
\begin{aligned} l^{*} &=(\left\lfloor\frac{s}{2}\right\rfloor+ x s)-x_{0}, \quad t^{*}=(\left\lfloor\frac{s}{2}\right\rfloor+ y s)-y_{0} \\ r^{*} &=x_{1}-(\left\lfloor\frac{s}{2}\right\rfloor+ x s), \quad b^{*}=y_{1}-(\left\lfloor\frac{s}{2}\right\rfloor+ y s) \end{aligned}
\end{equation}
where $\left(x_{0}, y_{0}\right)$ and $\left(x_{1}, y_{1}\right)$ denote
the left-top and right-bottom corners of the ground-truth bounding box $B^{*}$ associated with point $(x,y)$. 

Each location $(x,y)$ on the feature map of both classification and regression head, corresponds to an image patch on the input image centered at location $\left(\left\lfloor\frac{s}{2}\right\rfloor+ x s,\left\lfloor\frac{s}{2}\right\rfloor+ y s\right)$. Following \textbf{G2}, we directly classify corresponding image patch and regress the target bounding box at the location, as in many previous tracker~\cite{henriques2014high-speed,bolme2010visual,bertinetto2016fully}. In other words, our SiamFC++ directly views locations as training samples. While the anchor-based counterparts~\cite{li2018high,zhu2018distractor,li2019siamrpn++}, which consider the location on the input image as the center of multiple anchor boxes, output multiple classification score at the same location and regress the target bounding box with
respect to these anchor boxes, leading to ambiguous matching between anchor and object. Although \cite{li2018high,zhu2018distractor,li2019siamrpn++} have shown superior performance on various benmarks than \cite{henriques2014high-speed,bolme2010visual,bertinetto2016fully}, we empirically show that the ambiguous matching could result in serious issues (see Section \ref{subsec:anchor-based analysis} for more details).
In our per-pixel prediction fashion, only one prediction is made at each pixel on the final feature map. Hence it is clear that each classification score directly gives the confidence that the target is in the sub-window of the corresponding pixel and our design is free of ambiguity to this extent.

Since SiamFC++ does classification and regression w.r.t. the location, it is free of pre-defined anchor boxes, hence free of prior knowledge about target data distribution (e.g. scale/ratio), which comply with \textbf{G3}.

During the above sections, we do not take the target state estimation quality into consideration yet and directly use classification score to select the final box. That could cause the degradation of localization accuracy, as \cite{jiang2018acquisition} shows that classification confidence is not well correlated with the localization accuracy. According to the analysis in \cite{luo2016understanding}, input pixels around the center of a sub-window will have larger importance on the corresponding output feature pixel than the rest. Thus we hypothesize that feature pixels around the center of objects will have a better estimation quality than others. Following \textbf{G4}, we add a simple yet effective quality assessment branch similar to \cite{tian2019fcos,jiang2018acquisition} by appending a $1\times1$ convolution layer in parallel with the $1\times1$ convolution classification head, as shown in the right part of Figure \ref{fig:SiamFC++ pipeline}. The output is supposed to estimate the Prior Spatial Score (PSS) which is defined as follows:
\begin{equation}
    \mathrm{PSS}^{*} = \sqrt{\frac{\min(l^{*},r^{*})}{\max(l^{*},r^{*})} \times \frac{\min(t^{*},b^{*})}{\max(t^{*},b^{*})}}
\end{equation}
Note that PSS is not the only choice for quality assessment. As a variant, we can also predict the IoU score between predicted boxes and ground-truth boxes similar to \cite{jiang2018acquisition}:
\begin{equation}
    \mathrm{IoU}^{*} = \frac{\text{Intersection}(B,B^{*})}{\text{Union}(B,B^{*})}
\end{equation}
where $B$ is the predicted bounding box and $B^{*}$ is its corresponding ground-truth bounding box.

During inference, the score used for final box selection is computed by multiplying the PSS with the corresponding predicted classification score. In this way, those bounding boxes far from the center of objects will be downweighted seriously. Thus the tracking accuracy is improved. 

\subsection{Training Objective}
We optimize a training objective as follows:
\begin{equation}
\begin{split}
    L\left(\left\{{p}_{x, y}\right\}, {q}_{x, y}, \left\{\boldsymbol{t}_{x, y}\right\}\right) =\frac{1}{N_{\mathrm{pos}}} \sum_{x, y} &L_{\mathrm{cls}}\left({p}_{x, y}, c_{x, y}^{*}\right)\\
    +\frac{\lambda}{N_{\mathrm{pos}}} \sum_{x, y} 
    \mathbf{1}_{\left\{c_{x, y}^{*}>0\right\}}
    &L_{\mathrm{quality}}\left({q}_{x, y}, {q}_{x, y}^{*}\right) \\
    +\frac{\lambda}{N_{\mathrm{pos}}} \sum_{x, y} \mathbf{1}_{\left\{c_{x, y}^{*}>0\right\}} &L_{\mathrm{reg}}\left(\boldsymbol{t}_{x, y}, \boldsymbol{t}_{x, y}^{*}\right)
\end{split}
\end{equation} 
where $\mathbf{1}_{\{\cdot\}}$ is the indicator function that takes 1 if the condition in subscribe holds and takes 0 if not, $L_{\text{cls}}$ denote the focal loss~\cite{lin2017focal} for classification result, $L_{\text{quality}}$ denote the binary cross entropy (BCE) loss for quality assessment and $L_{\text{reg}}$ denote the IoU loss~\cite{yu2016unitbox} for bounding box result. We assign $1$ to $c_{x, y}^{*}$ if $(x,y)$ is considered as a positive sample, and $0$ if as a negative sample.

\newcounter{siamfc_ablat_cnt}
\begin{table*}
\tiny
\begin{center}
\tabcolsep=4.5pt
\resizebox{1.95\columnwidth}{!}{%
\begin{tabular}{|c|c|c|c|c|c|c|c|c|c|c|c|}
\hline
No. & VID &Youtube & COCO\&Det\& LaSOT\&GOT & Backbone & Head type & Head structure & Quality assessment & A & R & EAO & $\Delta$EAO \\
\hline
\hline
\refstepcounter{siamfc_ablat_cnt}\arabic{siamfc_ablat_cnt}
\label{ablat-siamfc-1} &
$\checkmark$ & $\times$ & $\times$ & AlexNet & cls & 0$\times$conv$3\times3$ & None & 0.506 & 0.566 & 0.213 & 0 \\
\refstepcounter{siamfc_ablat_cnt}\arabic{siamfc_ablat_cnt}
\label{ablat-siamfc-2} &
$\checkmark$ & $\checkmark$ & $\times$ & AlexNet & cls & 0$\times$conv$3\times3$ & None & 0.532	& 0.407 & 0.276 & +0.063 \\
\refstepcounter{siamfc_ablat_cnt}\arabic{siamfc_ablat_cnt}
\label{ablat-siamfc-3} &
$\checkmark$ & $\checkmark$ & $\times$ & AlexNet & cls & 3$\times$conv$3\times3$ & None & 0.539 & 0.337 & 0.296 & +0.083 \\
\refstepcounter{siamfc_ablat_cnt}\arabic{siamfc_ablat_cnt}
\label{ablat-siamfc-4} &
$\checkmark$ & $\checkmark$ & $\checkmark$ & AlexNet & cls & 3$\times$conv$3\times3$ & None & 0.536 & 0.323 & 0.306 & +0.093 \\


\refstepcounter{siamfc_ablat_cnt}\arabic{siamfc_ablat_cnt}
\label{ablat-siamfc-6} &
$\checkmark$ & $\checkmark$ & $\checkmark$ & AlexNet & cls+reg & 3$\times$conv$3\times3$ & PSS & 0.556 & \textbf{0.183} & 0.400 & +0.187 \\
\refstepcounter{siamfc_ablat_cnt}\arabic{siamfc_ablat_cnt}
\label{ablat-siamfc-7} &
$\checkmark$ & $\checkmark$ & $\checkmark$ & GoogLeNet & cls+reg & 2$\times$conv$3\times3$ & PSS & \textbf{0.587} & \textbf{0.183} & \textbf{0.426} & \textbf{+0.213} \\

\hline
\end{tabular}
}%
\end{center}
\caption{Ablation study: from SiamFC towards SiamFC++. Experiments have been conducted on VOT-2018 (A/R/EAO). $\Delta$EAO denotes the augmentation of EAO w.r.t. the baseline (Line \ref{ablat-siamfc-1}).}
\label{tab:ablation study:from SiamFC towards SiamFC++}
\end{table*}

\section{Experiments}
\label{sec:experiments}
\subsection{Implementation Details}

\paragraph{Model settings} 
In this work, we implement two versions of trackers with different backbone architectures: the one that adopts the modified version of AlexNet in the previous literature~\cite{bertinetto2016fully}, denoted as SiamFC++-AlexNet, and another one that uses GoogLeNet~\cite{szegedy2015going}, denoted as SiamFC++-GoogLeNet. With lower computation cost, the later achieves the same or even better performance(see Section \ref{sec:Results on different benchmarks}) on tracking benchmark than same previous methods using ResNet-50~\cite{he2016deep}. Both networks are pretrained on ImageNet~\cite{krizhevsky2012imagenet}, which has been proven practical for tracking task~\cite{li2018high,zhu2018distractor}. We will release the code to facilitate further researches.

\paragraph{Training data}
We adopt ILSVRC-VID/DET~\cite{russakovsky2015imagenet}, COCO~\cite{lin2014microsoft} , Youtube-BB~\cite{real2017youtube}, LaSOT~\cite{fan2019lasot} and GOT-10k~\cite{huang2018got} as our basic training set. Exceptions w.r.t. to specific benchmarks are detailed in the following subsections. For video datasets, we extract image pairs from VID, LaSOT, and GOT-10k by choosing frame pairs within an interval of less than 100 (5 for Youtube-BB). For image datasets (COCO/Imagenet-DET), we generate training samples by involving negative pairs~\cite{zhu2018distractor} as part of training samples to enhance the capacity to distinguish distractors of our model. We perform random shifting and scaling following a uniform distribution on the search image as data augmentation techniques. 

\paragraph{Training phase}
\label{parag:training phase}
For the AlexNet version, we freeze the parameters from conv1 to conv3 and fine-tune conv4 and conv5. For those layers without pretraining, we adopt a zero-centered Gaussian distribution with a standard deviation of 0.01 for initialization. We first train our model with for 5 warm up epochs with learning rate linearly increased from $10^{-7}$ to $2\times10^{-3}$, then use a cosine annealing learning rate schedule for the rest of 45 epochs, with 600k image pairs for each epoch. We choose stochastic gradient descent (SGD) with a momentum of 0.9 as the optimizer. 

For the version implemented with GoogLeNet, we freeze stage 1 and 2, fine-tune stage 3 and 4, augment the base learning rate to $2\times10^{-2}$, and multiply the learning rate of parameters in the backbone by 0.1 w.r.t the global learning rate. We also reduce the number of image pairs per epoch to 300k, reduce the total epoch to 20 (thus 5 for warming-up, and 15 for training) and unfreeze the parameters in backbone at the 10th epoch to avoid overfitting. For the experiment on LaSOT benchmark~\cite{fan2019lasot} (protocol II), we freeze the parameters in the backbone and further reduce the number of image pairs per epoch to 150k so that the training with the relatively smaller amount of training data could be stabilized.

The proposed tracker with AlexNet backbone runs at 160 FPS on the VOT2018 short-term benchmark, while the one with GoogleNet backbone runs at about 90 FPS on the VOT2018 short-term benchmark, both evaluated on an NVIDIA RTX 2080Ti GPU.

\paragraph{Test phase}
\label{parag:test phase}
The output of our model is a set of bounding boxes with their corresponding confidence scores $s$. Scores are penalized based on the scale/ratio change of corresponding boxes and distance away from the target position predicted in the last frame. Then the box with the highest penalized score is chosen and is used to update the target state.

\subsection{From SiamFC towards SiamFC++}
\label{ablation study}
While both employing a per-pixel prediction fashion, there exists a significant performance gap between SiamFC and our SiamFC++. In this subsection we perform an ablation study on VOT2018 dataset, with SiamFC as the baseline, aiming at identifying the key component for the improvement of tracking performance.

Results are shown in Table \ref{tab:ablation study:from SiamFC towards SiamFC++}. Concretely, in the SiamFC baseline, the tracker only performs classification tasks in its network and the target state estimation is done with multi-scale test. We gradually update SiamFC tracker by using extra training data (Line \ref{ablat-siamfc-2}/\ref{ablat-siamfc-4}), applying a better head structure (Line \ref{ablat-siamfc-3}), and adding the regression branch for accurate estimation to yield our proposed SiamFC++ tracker (Line \ref{ablat-siamfc-6}). We further replace the AlexNet backbone with GoogLeNet which is more powerful to extract visual feature (Line \ref{ablat-siamfc-7}).

The key components for tracking performance can be listed in descending order as follows: the regression branch (0.094), data source diversity (0.063/0.010), stronger backbone (0.026), and better head structure (0.020), where the $\Delta$EAO brought by each part is noted in parentheses. Note that these are the extra components of SiamRPN++ over SiamFC. After adding all the extra components into SiamFC, Our SiamFC++ achieves superior performance with less computation budget. Also, there are two things worth to mention: 1). the robustness (\textbf{R}) of Line \ref{ablat-siamfc-2} surpasses SiamRPN tracker ($0.46$~\cite{li2018high}); 2). the \textbf{R} of Line \ref{ablat-siamfc-3} is at the same level of DaSiamRPN ($0.337$~\cite{zhu2018distractor}) while using less data (without COCO and DET) than the latter. These results indicate that, while the introduction of the RPN module and anchor boxes setting undoubtedly gives better accuracy, its robustness is not improved and even hindered. We owe this to its violation of our proposed guidelines.

\paragraph{Quality Assessment Choice}
On GOT-10k \textit{val} subset, we obtain an AO of 77.8 for the tracker predicting PSS and an AO of 78.0 for the tracker predicting IoU. Experiments have been conducted with SiamFC++-GoogLeNet. We finally choose PSS in this paper as an implementation of our approach for its stability empirically observed across datasets during our experiment.

\subsection{Results on Several Benchmarks}
\label{sec:Results on different benchmarks}
We test our tracker on several benchmarks and results are gathered in Table \ref{tab:benchmark results}.

\subsubsection{Results on OTB2015 Benchmark}
As one of the most classical benchmarks for the object tracking task, the OTB benchmark~\cite{wu2013online} provides a fair test for all families of trackers. We conduct experiments on OTB2015~\cite{wu2013online} which contains 100 videos for tracker performance evaluation. With a success score of 0.682, our tracker reaches the state-of-the-art level w.r.t. other trackers in comparison. 

\subsubsection{Results on VOT Benchmark}
VOT2018~\cite{kristan2018sixth} contains 60 video sequences with several challenging topics including fast motion, occlusion, etc. We test our tracker on this benchmark and present the results in Table \ref{tab:benchmark results}. Both versions of our trackers reaching comparable scores w.r.t. current state-of-the-art trackers, the tracker with AlexNet backbone outperforms other trackers with the same tracking speed and while the tracker with GoogLeNet backbone yields a comparable score. Besides, our tracker has a significant advantage in the robustness among the trackers in comparison. To the best of our knowledge, this is the first tracker that achieves an EAO of 0.400 on VOT2018~\cite{kristan2018sixth} benchmark while running at a speed over 100 FPS, which demonstrate its potential of being applied in real production cases.

\subsubsection{Results on LaSOT Benchmark}
With a large number of video sequences (1400 sequences under Protocol I while 280 under Protocol II), LaSOT~\cite{fan2019lasot} (Large-scale Single Object Tracking) benchmark makes it impossible for trackers to overfit the benchmark, which achieves the purpose of testing the real performance of object tracking. Following Protocol II under which trackers are trained on LaSOT \textit{train} subset and evaluated on LaSOT \textit{test} subset, the proposed SiamFC++ tracker achieves better performance, even w.r.t. those who have better performance than ours on the VOT2018 benchmark. This reveals the fact that the scale of the benchmark influences the rank of trackers. 

\subsubsection{Results on GOT-10k Benchmark}
For target class generalization testing, we train and test our SiamFC++ model on GOT-10k~\cite{huang2018got} (Generic Object Tracking-10k) benchmark. Not only as a large-scale dataset (10,000 videos in train subset and 180 in both \textit{val} and \textit{test} subset), it also gives challenges in terms of the requirement of category-agnostic for generic object trackers as there is no class intersection between \textit{train} and \textit{test} subsets. We follow the protocol of GOT-10k and only trained our tracker on the \textit{train} subset. Our tracker with AlexNet backbone reaches an AO of 53.5 surpassing SiamRPN++ by 1.7, while our tracker with GoogLeNet backbone yields 59.5 which is even superior to ATOM that uses online updating method. This result shows the ability of our tracker to generalize even the target classes are unseen during the training phase, which matches the demand of the generic tracking.

\subsubsection{Results on TrackingNet Benchmark}
We evaluate our approach with 511 videos provided in the \textit{test} split of TrackingNet~\cite{muller2018trackingnet}. We exclude the Youtube-BB dataset from our training data in order to avoid data leak. As is described in \cite{muller2018trackingnet}, the evaluation server calculates the following three indexes based on tracking results: success rate, precision, and normalized precision. Our SiamFC++-GoogLeNet outperforms the current state-of-the-art methods (including online-update methods like \cite{danelljan2019atom}) in both precision and success rate dimensions, while our lightweight version SiamFC++ strikes a balance between performance and the speed. This result is achieved even without Youtube-BB containing a large portion of training data, which shows that the potential of our approach to be independent of large offline training data.

\begin{table}
\tiny
\begin{center}
\tabcolsep=.8pt
\resizebox{.95\columnwidth}{!}{ %
\begin{tabular}{c c|c c c c c c c}
\toprule
\multicolumn{2}{c|}{Trackers} & 
\begin{tabular}{c} SiamFC \\ \shortcite{bertinetto2016fully} \end{tabular} &
\begin{tabular}{c} ECO \\ \shortcite{danelljan2017eco} \end{tabular} &
\begin{tabular}{c} MDNet \\ \shortcite{nam2016learning} \end{tabular} &
\begin{tabular}{c} SiamRPN++ \\ \shortcite{li2019siamrpn++} \end{tabular} &
\begin{tabular}{c} ATOM \\ \shortcite{danelljan2019atom} \end{tabular} &
\begin{tabular}{c} \textbf{SiamFC++-} \\ \textbf{AlexNet} \end{tabular}  &
\begin{tabular}{c} \textbf{SiamFC++-} \\ \textbf{GoogLeNet} \end{tabular} \\

\midrule
\multirow{1}{*}{OTB-15} 
& Success & 58.2 & {\color{red}70.0} & 67.8 & {\color{green}69.6} & 66.9 & 65.6 & {\color{blue}68.3} \\
\midrule
\multirow{3}{*}{VOT-18}
& A   & 0.503 & 0.484 & - & {\color{red}0.600} & {\color{green}0.590} & 0.556 & {\color{blue}0.587} \\
& R   & 0.585 & 0.276 & - & 0.234 & {\color{blue}0.204} & {\color{red}0.183} & {\color{red}0.183} \\
& EAO & 0.188 & 0.280 & - & {\color{green}0.414} & {\color{blue}0.401} & 0.400 & {\color{red}0.426} \\
\midrule
\multirow{1}{*}{LaSOT} 
& Success & 33.6 & 32.4 & 39.7 & 49.6 & {\color{green}51.5} & {\color{blue}50.1} & {\color{red}54.4} \\
\midrule
\multirow{3}{*}{GOT} 
& SR\textsubscript{.5} & 35.3 & 30.9 & 30.3 & {\color{blue}61.8} & {\color{green}63.4} & 57.7 & {\color{red}69.5} \\
& SR\textsubscript{.75} & 9.8 & 11.1 & 9.9 & {\color{blue}32.5} & {\color{green}40.2} & 32.3 & {\color{red}47.9} \\
& AO & 34.8 & 31.6 & 29.9 & {\color{blue}51.8} & {\color{green}55.6} & 49.3 & {\color{red}59.5} \\
\midrule
\multirow{3}{*}{T-Net} 
& Prec.       & 51.8 & 49.2 & 56.5 & {\color{green}69.4} & {\color{blue}64.8} & 64.6 & {\color{red}70.5} \\
& Norm. Prec. & 65.2 & 61.8 & 70.5 & {\color{red}80.0} & {\color{blue}77.1} & 75.8 & {\color{red}80.0}\\
& Succ.     & 55.9 & 55.4 & 60.6 & {\color{green}73.3} & 70.3 & {\color{blue}71.2} & {\color{red}75.4}\\
\midrule
\multicolumn{2}{c|}{FPS} & {\color{blue}86} & 8 & 1 & 35 & 30 & {\color{red}160} & {\color{green}90} \\
\bottomrule
\end{tabular}
}%
\end{center}
\caption{Results on several benchmarks. T-Net denotes TrackingNet. Top-3 results of each dimension (row) are colored in red, green, and blue, respectively.}
\label{tab:benchmark results}
\end{table}

\subsection{Comparison with Trackers that Do not Apply Our Guidelines}
\label{subsec:anchor-based analysis}

The family of SiamRPN~\cite{li2018high,zhu2018distractor,li2019siamrpn++} has achieved great success in visual tracking these years and drawn much attention from tracking community. Here we use state-of-the-art SiamRPN++ tracker as an example. Despite recent successes of the SiamRPN family, we have found that the SiamRPN tracker and its family do not follow our proposed guidelines entirely.
\begin{itemize}
    \item \textbf{(G2)} the classification score of SiamRPN represents the similarity between anchor and object, rather than template object and objects in search image, which may cause matching ambiguity;
    \item \textbf{(G3)} the design of pre-set anchor boxes needs prior knowledge of the distribution of size and ratio of target;
    \item \textbf{(G4)} the choice of target state estimation does not take estimation quality into consideration.
\end{itemize}

Note that the SiamRPN family adopts proposal refinement by the regression branch instead of the multi-scale test, and thus achieves astonishing tracking accuracy, which complies with our guideline \textbf{G1}. 

As a consequence of the violation of guideline \textbf{G2}, we empirically find that the SiamRPN family is prone to deliver a false-positive result. In other words, SiamRPN will produce an unreasonable high score for nearby objects or background under large appearance variation of target object. As shown in Figure \ref{fig:comparison anchor-based/anchor-free}, we can see that SiamRPN++ fails to track the target object by giving very high scores for nearby objects (i.e. a rock or a face) under challenging scenarios like out-of-plane rotation and deformation. We hypothesize that SiamRPN matches objects and anchors rather than the object itself, which may deliver drifts and thus hinders its robustness. On the contrary, our proposed SiamFC++, which matches between template objects and objects in search image directly, gives accurate score predictions and successfully tracks the target. 

\begin{figure}[t] 
    \centering
    \includegraphics[width=.95\columnwidth]{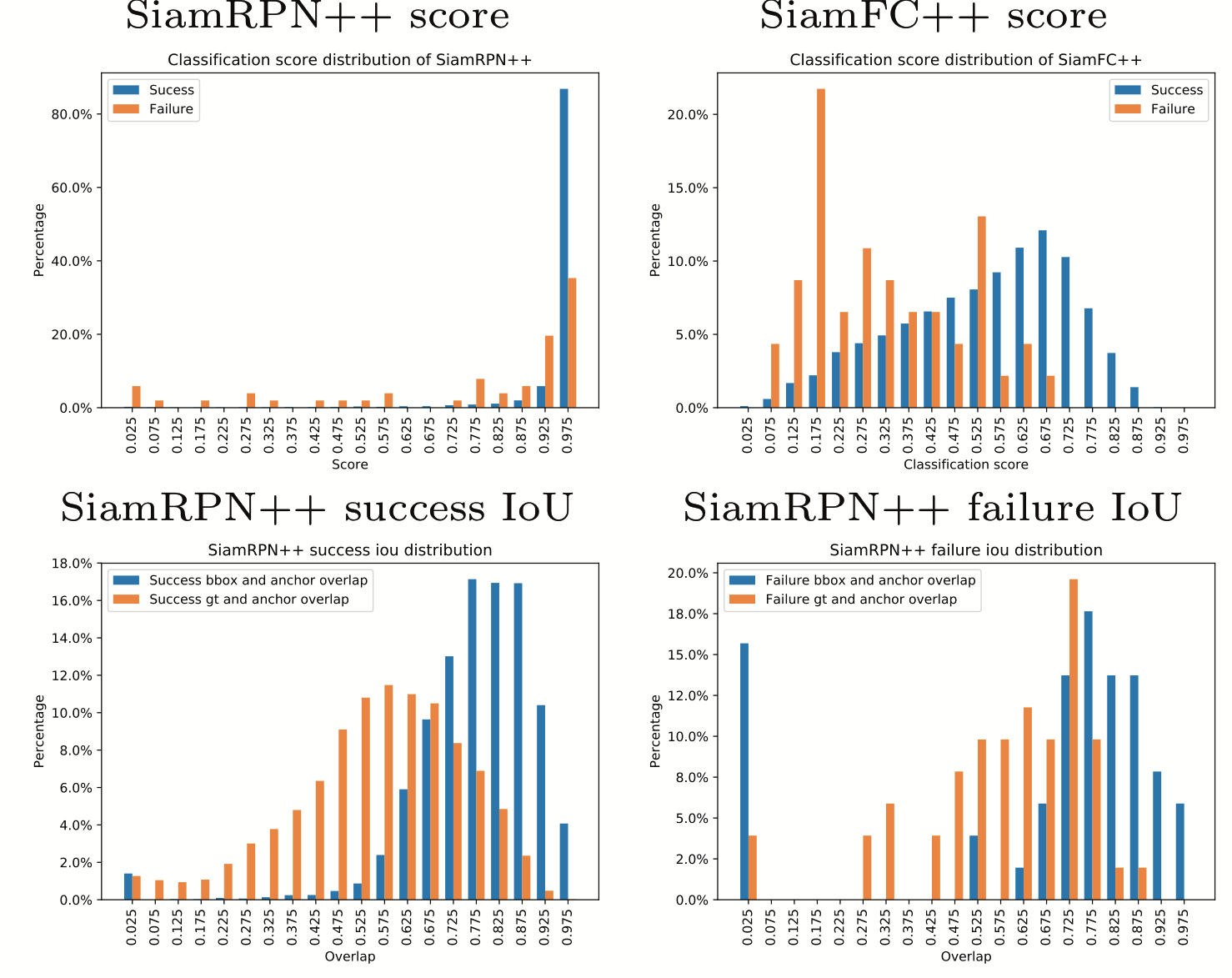}
    \caption{The first row: score distribution of SiamRPN++ and SiamFC++. The second row: IoU distribution of SiamRPN++ in both success/failure state. Better visualized when zoomed in.}
    \label{fig:score/iou distribution}
\end{figure}

To verify our hypothesis, we record the max score produced by SiamRPN++ and our proposed SiamFC++ on VOT2018 dataset. We then split them according to the tracking result, e.g., successful or failed. On VOT2018, a tracking result is considered failed if its overlap with the ground-truth box is zero. Otherwise, it is considered successful. The result is visualized in the first row in Figure \ref{fig:score/iou distribution}. Comparing SiamRPN++ and SiamFC++ scores, we can see that most classification score of SiamRPN++ follows similar and highly overlapped distributions, successful or not, while the classification score of our SiamFC++ of failure state exhibit very different pattern with that of a successful state. 

Another factor contributing to the ambiguity in SiamRPN++ is that the feature matching process is done with patches of fixed aspect ratio (multiple patches with different ratios will bring non-negligible computation cost), while each pixel of the feature after matching is assigned anchors whose aspect ratio varies.

As for the violation of \textbf{G3}, the performance of SiamRPN varies as the scales and ratios of anchors vary. As is shown in Table 3 from \cite{li2018high}, three different ratio settings are tried and the performance of SiamRPN varies when using different anchor settings. Thus the best performance is achieved only by accessing prior knowledge of data distribution, which is against the spirit of generic object tracking~\cite{huang2018got}.

Besides, in the second row of Figure \ref{fig:score/iou distribution}, we also plot the histogram of SiamRPN++ statistics of IoU between output bounding box and ground truth and the histogram between anchor and ground truth, in both success and failure state. As is shown from the IoU distribution, the prior knowledge given by anchor settings (violation of \textbf{G3}) leads to a bias in target state estimation. Concretely, the predicted box of SiamRPN++ tends to overlap more with the anchor box than with the ground truth box which can lead to performance degradation. 

As for the violation of \textbf{G4}, we can see that the SR\textsubscript{.5} and SR\textsubscript{.75} of SiamRPN++ on GOT-10k benchmark are 7.7 and 15.4 points lower than those of SiamFC++, respectively. In GOT-10k, the Success Rate (SR) measures the percentage of successfully tracked frames where the overlaps exceed a pre-defined threshold (i.e., 0.5 or 0.75). The higher the threshold, the more accurate the tracking result. Hence SR is a solid indicator for estimation quality. The SR\textsubscript{.75} of SiamRPN++ is much lower than that of SiamFC++, indicating the lower estimation quality of SiamRPN++ caused by the violation of guideline \textbf{G4}.

\section{Conclusion}
In this paper, we propose a set of guidelines for target state estimation in tracker design, by analyzing the unique characteristics of visual tracking tasks and the flaws of former trackers. Following these guidelines, we propose our approach that provides effective methods for both classification and target state estimation (\textbf{G1}), giving classification score without ambiguity (\textbf{G2}), tracking without prior knowledge (\textbf{G3}), and being aware of estimation quality (\textbf{G4}). We verify the effectiveness of proposed guidelines by extensive ablation study. And we show that our tracker based on these guidelines reaches state-of-the-art performance on five challenging benchmarks, while still running at 90 FPS.

{
\fontsize{9.0pt}{10.0pt}\selectfont
\bibliographystyle{aaai}
\bibliography{AAAI-XuY.5104}}

\clearpage
\begin{appendices}

\section{Backbone Choice}
Although with smaller MACs and parameter amounts (see Table \ref{tab:Backbone Stats} for statistics), our proposed tracker (version with GoogLeNet) achieves the same or even better performance of previous trackers using ResNet-50 as backbone across different benchmarks. On the other hand, this choice contributes significantly to the efficiency which is crucial for visual tracking task.

\section{Test Phase Behavior}
\label{sec:test phase behavior}
The output of our model is a set of bounding boxes $B$ with their corresponding confidence scores $s$ in the shape of $N \times N \times 4$ and $N \times N $, respectively. Following the tracking strategy of the SiamRPN family ~\cite{bertinetto2016fully,li2018high,zhu2018distractor}, we adopt the following strategies to render our tracking result more robust. 
\begin{itemize}
    \item First, the score map $s$ output by the model is multiplied by a penalization term $p$ according to ratio($r$ and $r^\prime$) and size changes($s$ and $s^\prime$) of their corresponding boxes. Namely,
    \begin{equation}
    \begin{split}
        p &=e^{k \cdot \max \left(\frac{r}{r^{\prime}}, \frac{r^{\prime}}{r}\right) \odot \max \left(\frac{s}{s^{\prime}}, \frac{s^{\prime}}{s}\right)}
    \\
        \overline{s} &= s \odot p
    \end{split}
    \end{equation}
    where $k$ is a hyperparameter that controls the magnitude of the penalization, $\odot$ denotes element-wise multiplication since the ratio($r$ and $r^\prime$) and the size changes($s$ and $s^\prime$) are all matrices in the shape of $N \times N$ (refer to ~\cite{li2018high} for more details).
    
    \item Second, a cosine window $\omega$ is added with a window influence coefficient $\Omega$ to suppress large displacement since smooth motion is assumed.
    \begin{equation}
    \begin{split}
        \omega(x) &= 0.5-0.5 \cos \left(\frac{2 \pi \cdot \text{dist} (x, x_c) }{(N-1)/2-1}\right)
    \\
        \tilde{s} &= \overline{s} \cdot (1- \Omega) + \omega \cdot \Omega
    \end{split}
    \end{equation}
    where $\Omega$ is a hyperparameter that controls the influence of window, $x$ is a point coordinate on the score map, $x_c$ is the center point coordinate on the score map, and $\text{dist}(i, j)$ denotes the Euclidean distance between the point $i$ and the point $j$. 
    
    \item Third, the point with the highest score on $\tilde{s}$ is chosen as $x^*$ and its corresponding bounding box $B[x^*]$ is selected as the estimation $B_\text{curr}$ for the current frame.
    \begin{equation}
    \begin{split}
        x^* &= \mathop{\arg\max}_{x\in[0..N-1]^{\otimes2}} \tilde{s}[x]
    \\
        B_{\text{curr}} &= B[x^*]
    \end{split}
    \end{equation}
    where $[0..N-1]^{\otimes2}$ refers to the coordinate space of the score map.
    
    \item Finally, the target size is updated by linear interpolation to make the shape change smoothly.
    \begin{equation}
    \begin{split}
        \alpha^\prime &= \overline{s}[x^*] \cdot \alpha
    \\
        B_{\text{pred}}.\text{size} &= (1 - \alpha^\prime) \cdot B_{\text{prev}}.\text{size} + \alpha^\prime \cdot B_{\text{curr}}.\text{size}
    \end{split}
    \end{equation}
    where $B_\text{prev}$ and $B_\text{pred}$ denotes the previous prediction and final prediction of target bounding box, respectively. Here $\alpha$ is a hyperparameter to control the target size updating speed, multiplied by the penalized score on $x^*$ to make the box size updating speed match its confidence. The $.\text{size}$ operation denotes the indexing on size dimensions(i.e. width and height).
\end{itemize}

\begin{table}[t]
\tiny
\begin{center}
\tabcolsep=6pt
\begin{tabular}{c|c c c c}
\toprule
Backbone & ResNet-50 & GoogleNet & ResNet-50 & GoogleNet \\
Application & ImageNet-CLS & ImageNet-CLS & SiamRPN++ & SiamFC++ \\
\midrule
Input size & $224 \times 224$ & $224 \times 224$ & $255 \times 255$ & $303 \times 303$ \\
MACs[G] & 4.14 & 2.9 & 44.7 & 15.9 \\
Parameters[M] & 25.6 & 27.2 & 45.5 & 9.16 \\
\bottomrule
\end{tabular}
\end{center}
\caption{Backbone characteristics. Input sizes for trackers indicated here are the real sizes during the test phase. In SiamFC++, the last stage of GoogLeNet have been pruned which results in a smaller parameter amount. }
\label{tab:Backbone Stats}
\end{table}

\section{Ablation Study over SiamFC++}

\newcounter{siamfc++_ablat_cnt}
\begin{table*}
\tiny
\begin{center}
\tabcolsep=4.5pt
\begin{tabular} {|c|c|c|c|c|c|c|c|c|c|c|c|c|c|}
\hline
No. & VID\&Youtube & COCO\&Det & LaSOT & GOT & Backbone & Head type & Head structure & Quality assessment & A & R & EAO & SR\textsubscript{.5} & AO \\
\hline
\hline

\refstepcounter{siamfc++_ablat_cnt}\arabic{siamfc++_ablat_cnt}
\label{ablat-data-1} &
$\checkmark$ & $\times$ & $\times$ & $\times$ & AlexNet & cls+reg & 3$\times$conv$3\times3$ & PSS & 0.561 & 0.258 & 0.352 & - & - \\
\refstepcounter{siamfc++_ablat_cnt}\arabic{siamfc++_ablat_cnt}
\label{ablat-data-2} &
$\checkmark$ & $\checkmark$ & $\times$ & $\times$ & AlexNet & cls+reg & 3$\times$conv$3\times3$ & PSS & 0.548 & 0.215 & 0.378 & - & - \\
\refstepcounter{siamfc++_ablat_cnt}\arabic{siamfc++_ablat_cnt}
\label{ablat-data-3}\label{ablat-basemodel-alex-vot} &
$\checkmark$ & $\checkmark$ & $\checkmark$ & $\checkmark$ & AlexNet & cls+reg & 3$\times$conv$3\times3$ & PSS & 0.556 & \textbf{0.183} & 0.400 & - & - \\
\refstepcounter{siamfc++_ablat_cnt}\arabic{siamfc++_ablat_cnt}
\label{ablat-basemodel-google-vot} &
$\checkmark$ & $\checkmark$ & $\checkmark$ & $\checkmark$ & GoogLeNet & cls+reg & 2$\times$conv$3\times3$ & PSS & \textbf{0.587} & \textbf{0.183} & \textbf{0.426} & - & - \\

\hline 

\refstepcounter{siamfc++_ablat_cnt}\arabic{siamfc++_ablat_cnt}
\label{ablat-head-1conv} &
$\times$ & $\times$ & $\times$ & $\checkmark$ & GoogLeNet & cls+reg & 1$\times$conv$3\times3$ & PSS & - & - & - & 88.4 & 76.2 \\
\refstepcounter{siamfc++_ablat_cnt}\arabic{siamfc++_ablat_cnt}
\label{ablat-head-2conv}\label{ablat-basemodel-google-got}
&
$\times$ & $\times$ & $\times$ & $\checkmark$ & GoogLeNet & cls+reg & 2$\times$conv$3\times3$ & PSS & - & - & - & 88.8 & 77.8 \\
\refstepcounter{siamfc++_ablat_cnt}\arabic{siamfc++_ablat_cnt}
\label{ablat-head-2conv-noctr} &
$\times$ & $\times$ & $\times$ & $\checkmark$ & GoogLeNet & cls+reg & 
2$\times$conv$3\times3$ & None & - & - & - & 88.6 & 77.1 \\
\refstepcounter{siamfc++_ablat_cnt}\arabic{siamfc++_ablat_cnt}
\label{ablat-head-2conv-iou_score} &
$\times$ & $\times$ & $\times$ & $\checkmark$ & GoogLeNet & cls+reg & 2$\times$conv$3\times3$ & IoU & - & - & - & 89.3 & 78.0 \\
\refstepcounter{siamfc++_ablat_cnt}\arabic{siamfc++_ablat_cnt}
\label{ablat-head-3conv} &
$\times$ & $\times$ & $\times$ & $\checkmark$ & GoogLeNet & cls+reg & 3$\times$conv$3\times3$ & PSS & - & - & - & \textbf{90.0} & \textbf{78.1} \\
\refstepcounter{siamfc++_ablat_cnt}\arabic{siamfc++_ablat_cnt}
\label{ablat-basemodel-alex-got} &
$\times$ & $\times$ & $\times$ & $\checkmark$ & AlexNet & cls+reg & 3$\times$conv$3\times3$ & PSS & - & - & - & 87.2 & 74.9 \\
\hline
\end{tabular}
\end{center}
\caption{Ablation study: better understanding of SiamFC++. Experiments have been conducted on VOT-2018(A/R/EAO) and GOT-10k val(SR\textsubscript{.5}/AO)}
\label{tab:ablation study:better understanding of SiamFC++}
\end{table*}

\paragraph{Data source diversity}
We conduct our ablation study on the diversity of data sources to identify the influence of data diversity on tracker performance. We choose three different training data settings varying from low to high diversity. The results in Line \ref{ablat-data-1}, \ref{ablat-data-2}, and \ref{ablat-data-3} of Table \ref{tab:ablation study:better understanding of SiamFC++} show that the augmentation of the diversity of training data significantly brings higher performance, which demonstrates the ability of our method to exploit rich offline training data.

\paragraph{Backbone capacity}
To unveil the power of visual representation brought by deep CNN networks, we conduct ablation experiments by replacing the backbone by a larger base model, GoogLeNet-v2~\cite{szegedy2015going,Ioffe:2015:BNA:3045118.3045167}. We adopt the technique of cropping the border of the backbone output feature map with a width of 4, which has been introduced in \cite{li2019siamrpn++} for deep CNN architecture in the tracking task. The results in Table \ref{tab:ablation study:better understanding of SiamFC++} show that the tracker with GoogLeNet outperforms the one with AlexNet  by 0.017 on VOT2018 (Line \ref{ablat-basemodel-alex-vot} and \ref{ablat-basemodel-google-vot}) and 3.2 on GOT-10k (Line \ref{ablat-basemodel-alex-got} and \ref{ablat-basemodel-google-got}), which indicates that a stronger backbone brings a better tracking result. Additionally, the gain in tracking performance is mainly brought by the enhancement of accuracy. 

\paragraph{Structure of head}
Due to the fact that padding may cause the drop of tracking performance by introducing noise, previous RPN-based siamese trackers~\cite{li2018high,zhu2018distractor,li2019siamrpn++} do not use conv$3\times3$ layer after the cross-correlation operation in order to avoid the shrinkage of response region(the region on the search image which has a corresponding point on the score map) which may hinder the tracking performance, particularly in case of fast motion.

However, in our SiamFC++ design, a trade-off can be achieved between the real search region size and the head capacity. An ablation study on the head structure is conducted and the results are shown in Line \ref{ablat-head-1conv}, \ref{ablat-head-2conv}, and \ref{ablat-head-3conv} of Table \ref{tab:ablation study:better understanding of SiamFC++}. The input size of the search image is fixed at $303 \times 303$ to ensure that the trackers see a search region of the same size for a fair comparison. The results show that the performance gets better as we add more conv$3\times3$ layers in the RPN head(0.4 brought by the second conv$3\times3$ layer and 0.3 brought by the third one) despite the shrinkage of response region. This proves that the gain from the enhancement of head capacity is superior to the gain from the increment of the response region range. In addition, the qualitative result in Figure \ref{fig:head struct qualitative} also reveals that even with a small part of the target object falling in the score map covered zone, the model is able to predict the correct classification/regression response. Moreover, the enhanced robustness to pose change, occlusion, and the presence of distractors is also shown in the qualitative result. 

Finally, we choose the tracker with two conv$3\times3$ layers in the head for it strikes a balance between the performance and the speed.

\def\SubFigureWidth{0.4\linewidth}
\begin{figure}[t]
  \centering
  \begin{minipage}[b]{\SubFigureWidth}
    \centering
    \includegraphics[width=.98\linewidth]{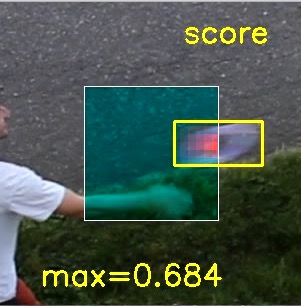}\par
    {fast motion}
  \end{minipage}
  \begin{minipage}[b]{\SubFigureWidth}
    \centering
    \includegraphics[width=1\linewidth]{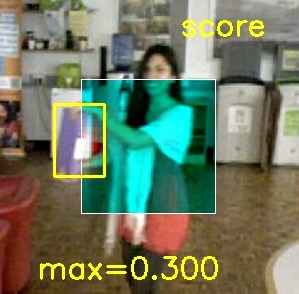}\par
    {fast motion}
  \end{minipage}

  \quad

  \begin{minipage}[b]{\SubFigureWidth}
    \centering
    \includegraphics[width=1\linewidth]{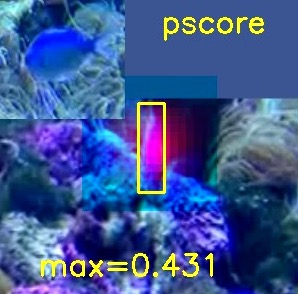}\par
    {pose change} 
  \end{minipage}
  \begin{minipage}[b]{\SubFigureWidth}
    \centering
    \includegraphics[width=1\linewidth]{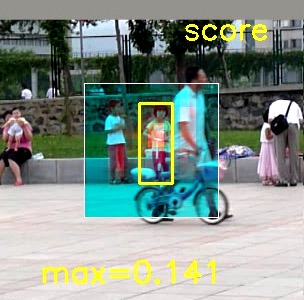}\par
    {occlusion}
  \end{minipage}
  \caption{Tracking result by our tracker with multiple conv$3\times3$ layers in the head. The first two images show its performance for fast-motion objects, the third image presents its capacity to deal with pose change, and the fourth image demonstrates the robustness in case of occlusion and the presence of distractors.}
  \label{fig:head struct qualitative}
\end{figure}

\def\StatW{.22\linewidth}
\def\StatH{.15\linewidth}
\begin{table*}[t]
\centering
\begin{tabular}{c c c c}
    \includegraphics[width=\StatW,height=\StatH]{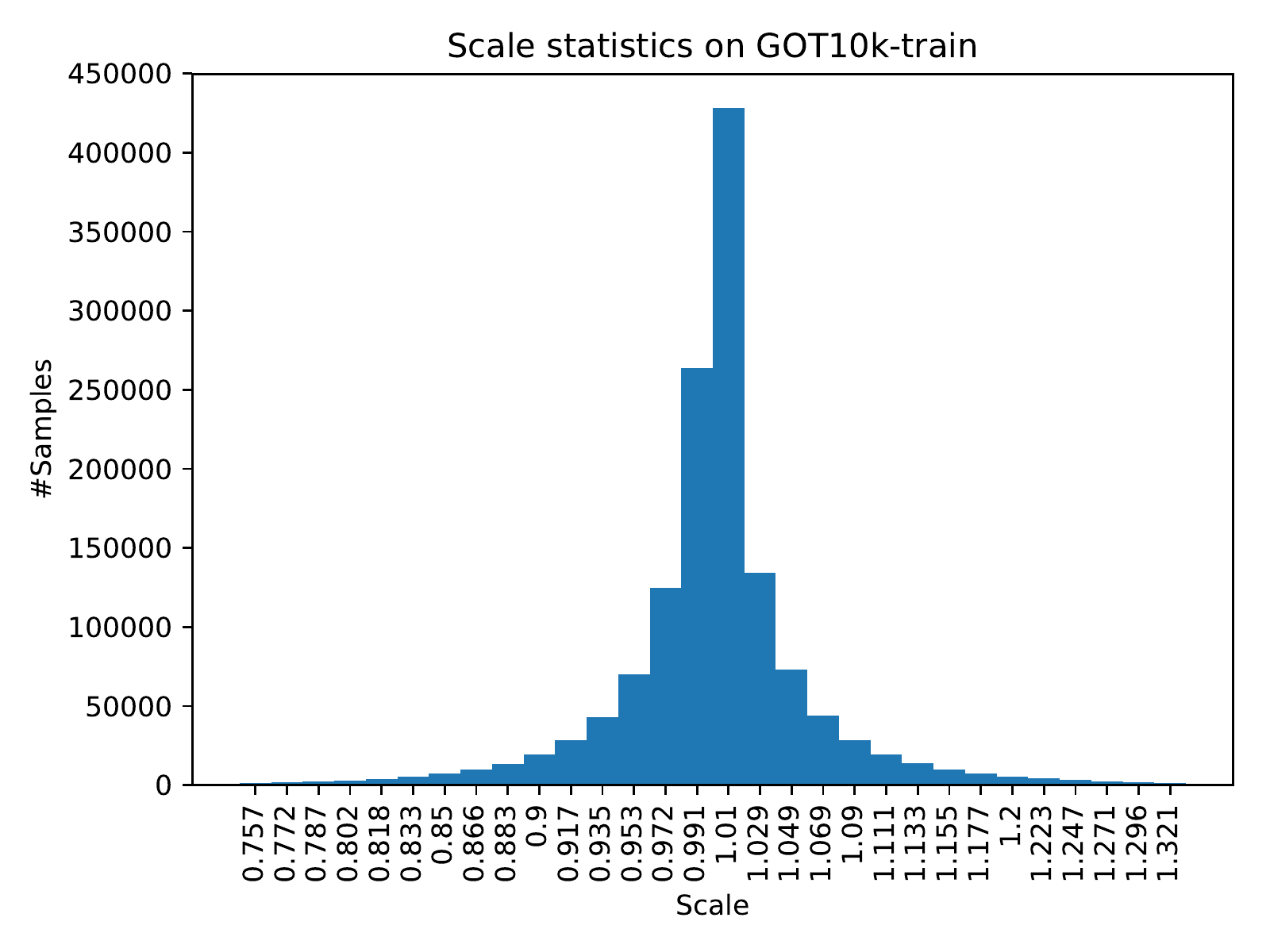}
    &
    \includegraphics[width=\StatW,height=\StatH]{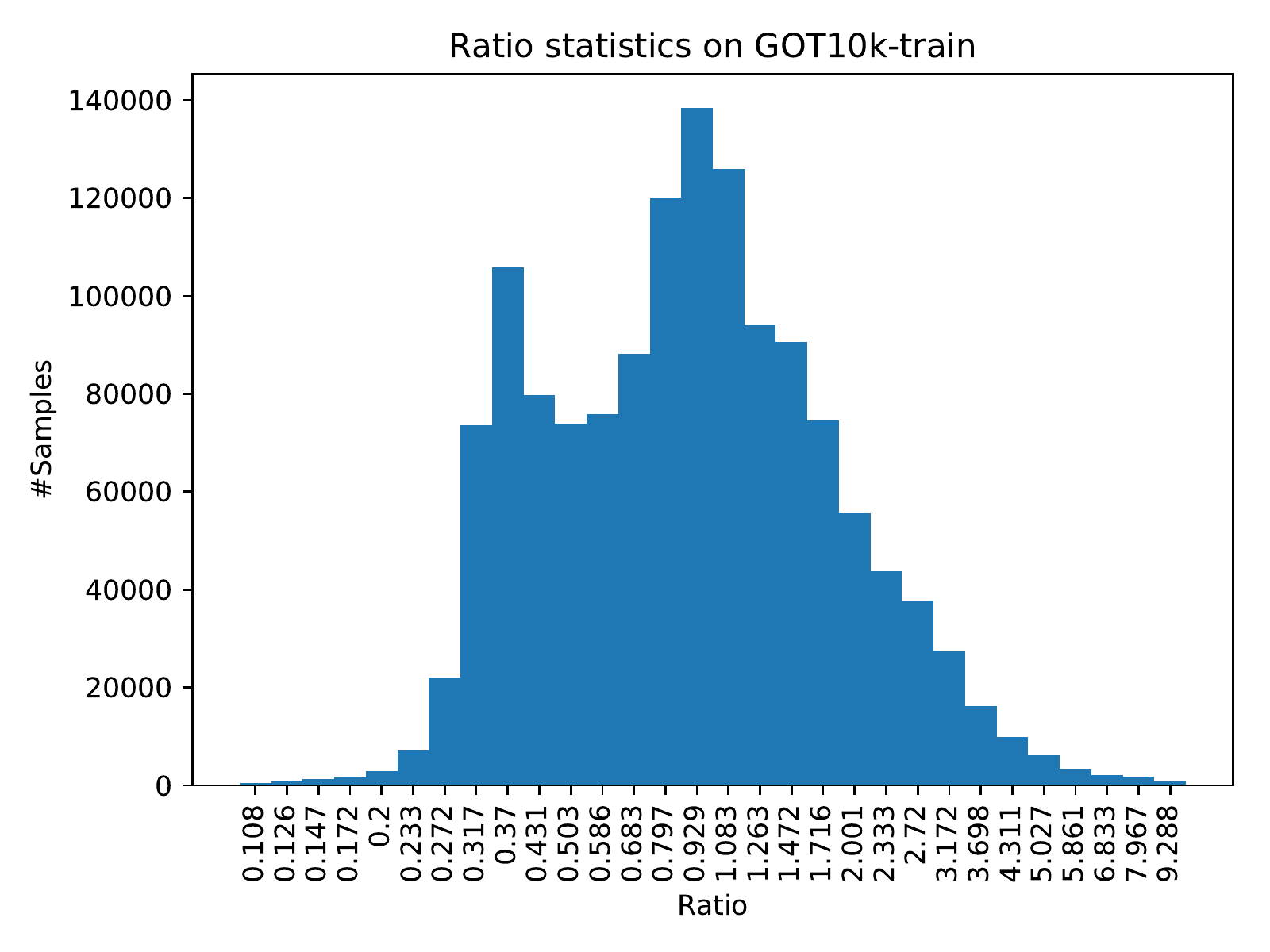}
    &
    \includegraphics[width=\StatW,height=\StatH]{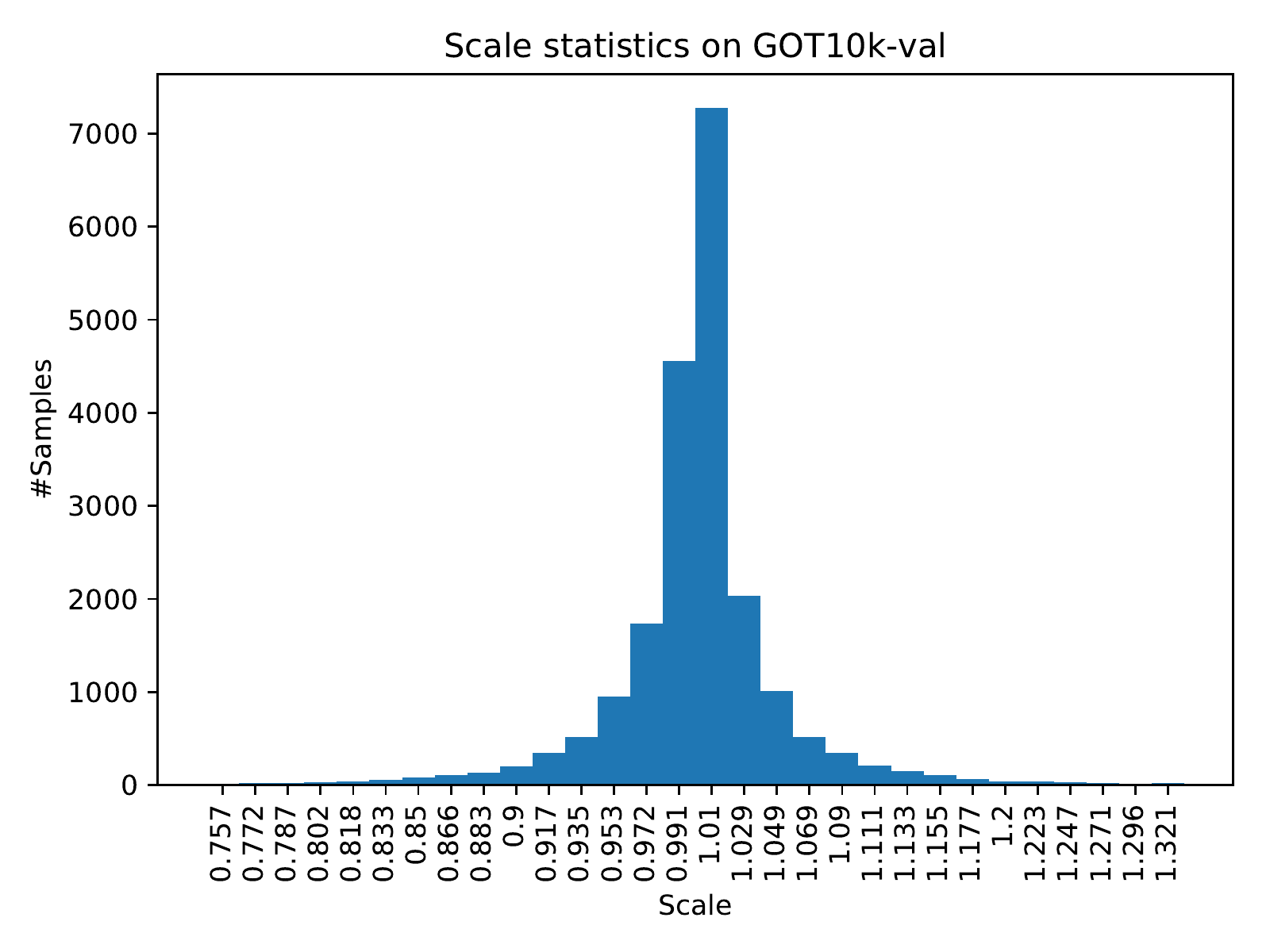}
    &
    \includegraphics[width=\StatW,height=\StatH]{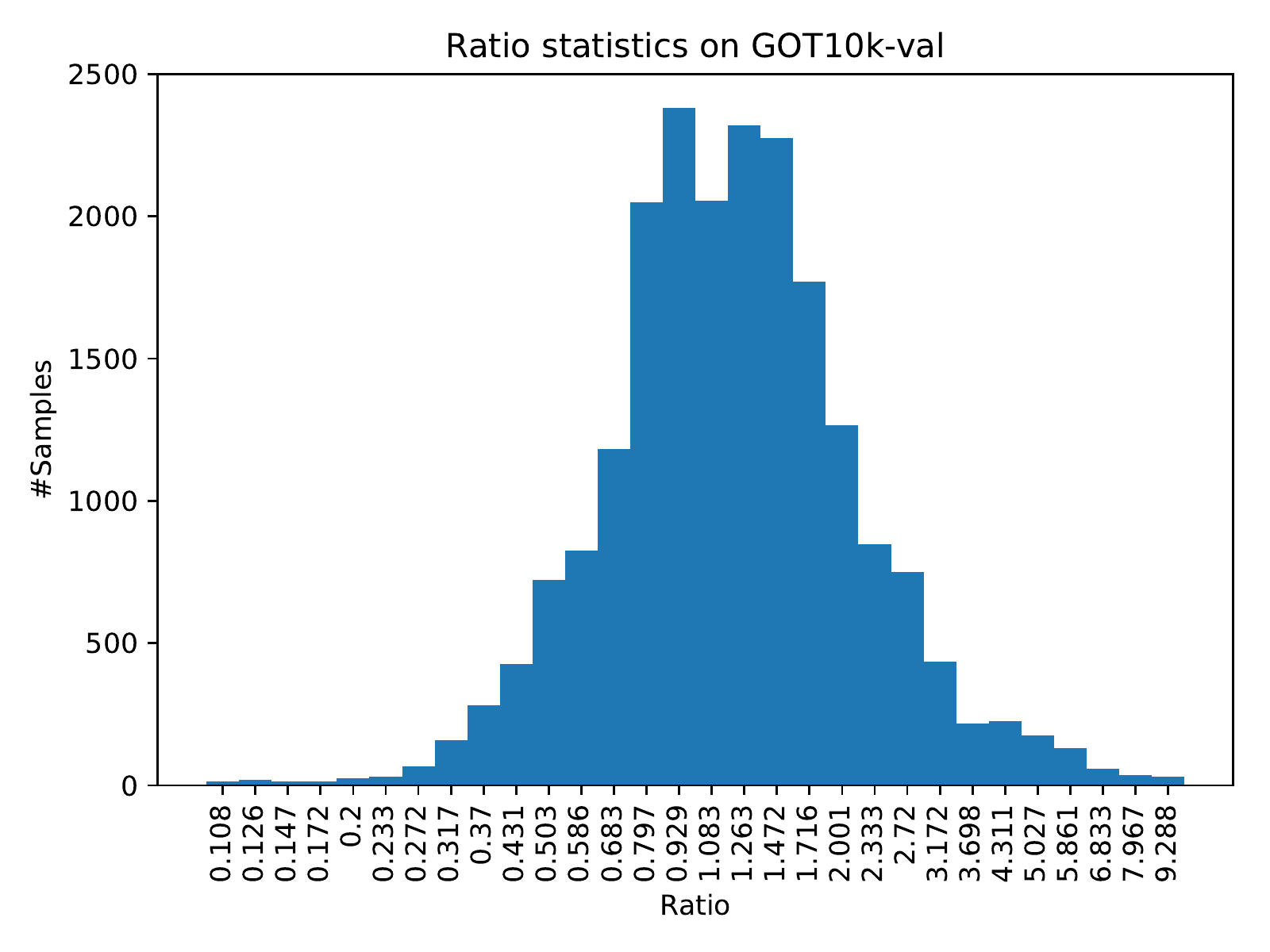}
\\
    \includegraphics[width=\StatW,height=\StatH]{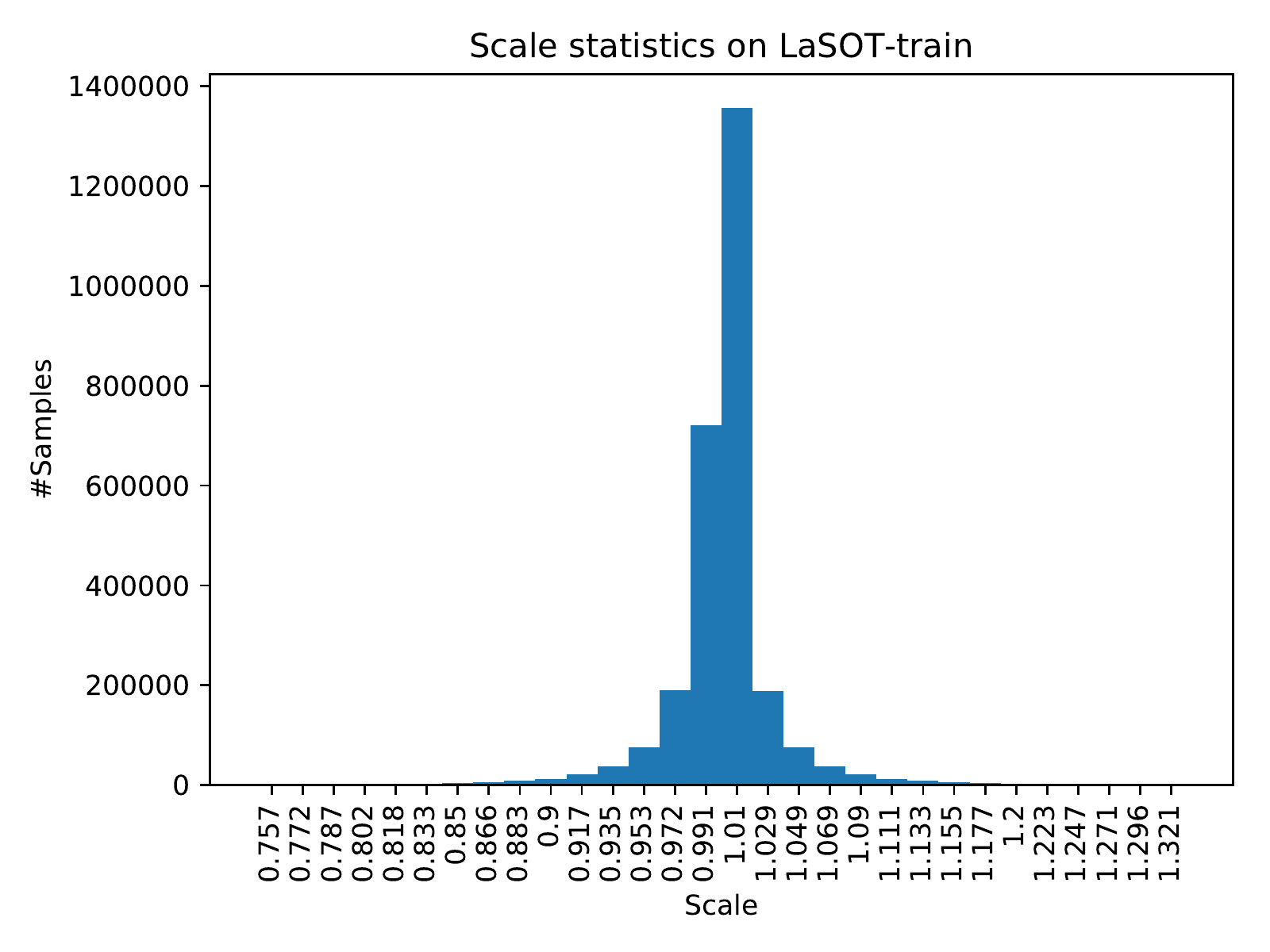}
    &
    \includegraphics[width=\StatW,height=\StatH]{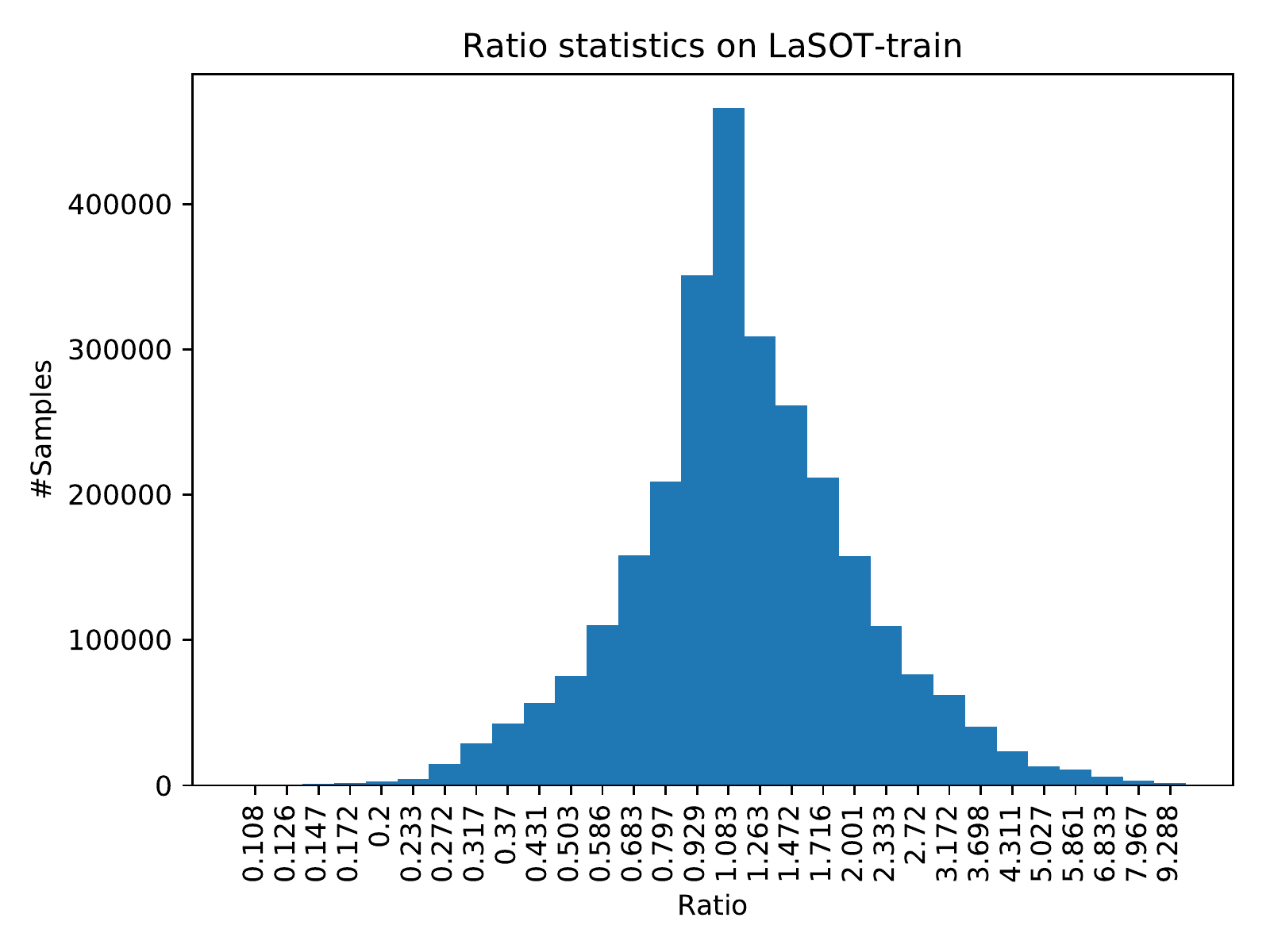}
    &
    \includegraphics[width=\StatW,height=\StatH]{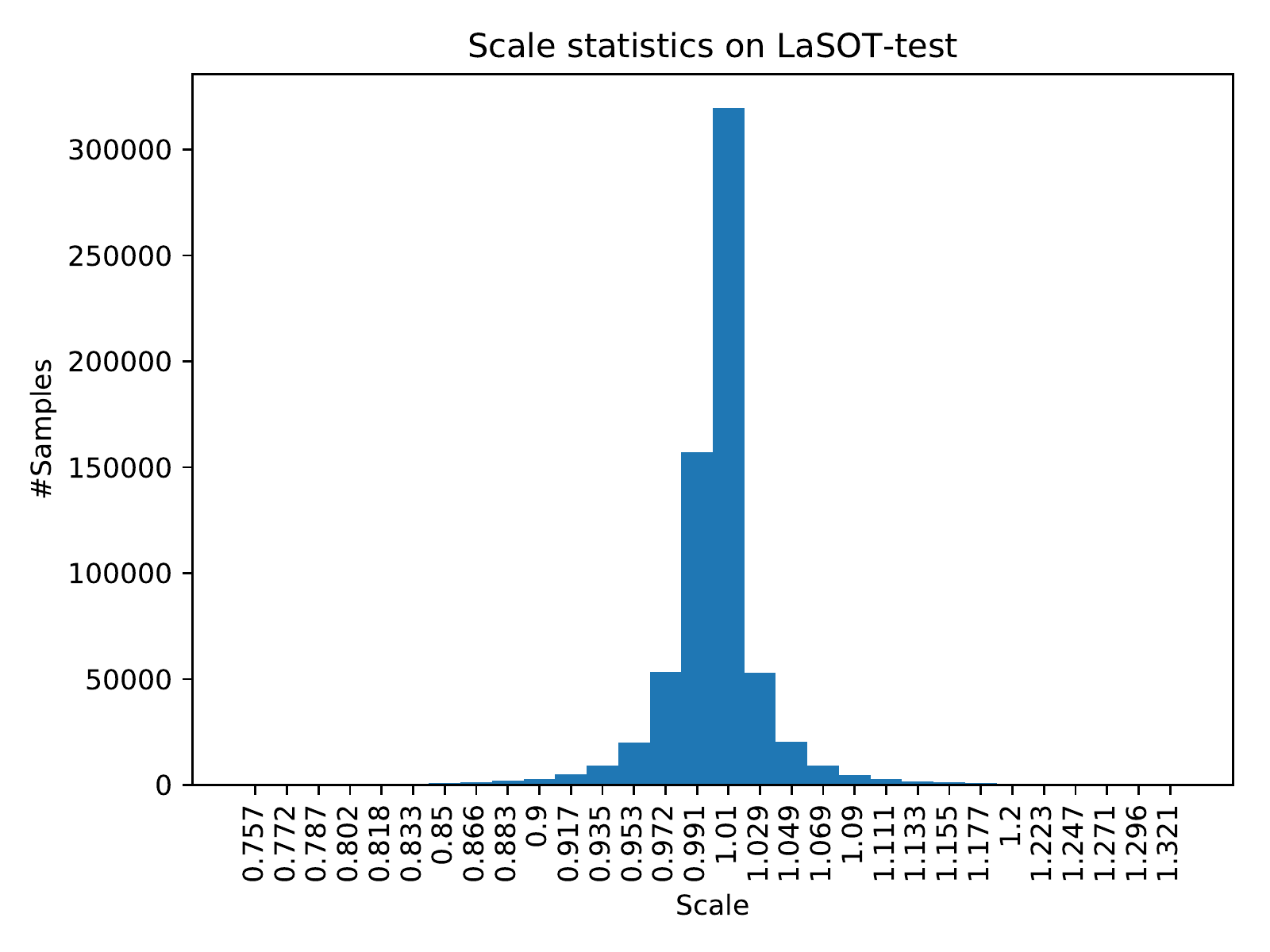}
    &
    \includegraphics[width=\StatW,height=\StatH]{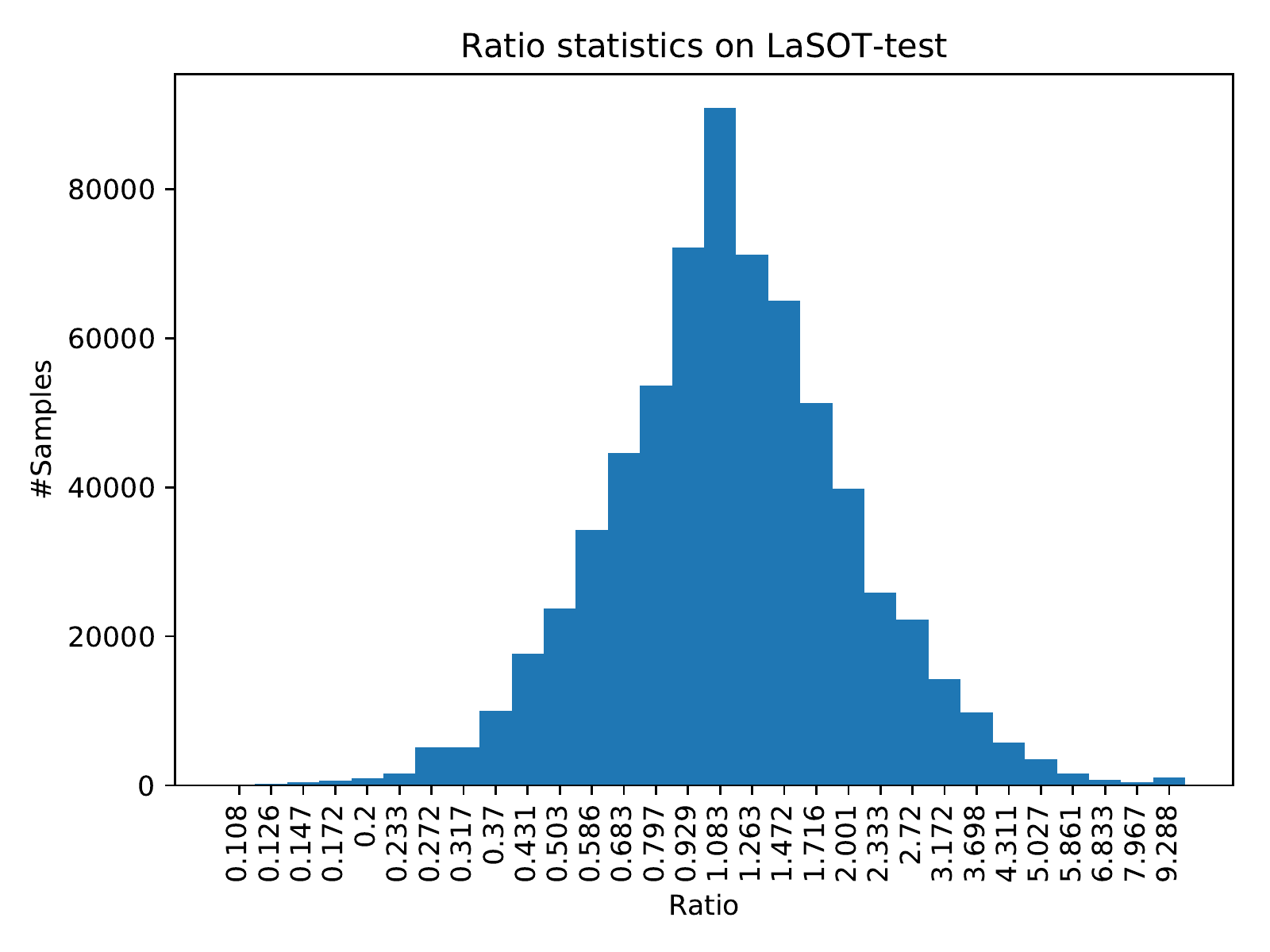}
\\
    \includegraphics[width=\StatW,height=\StatH]{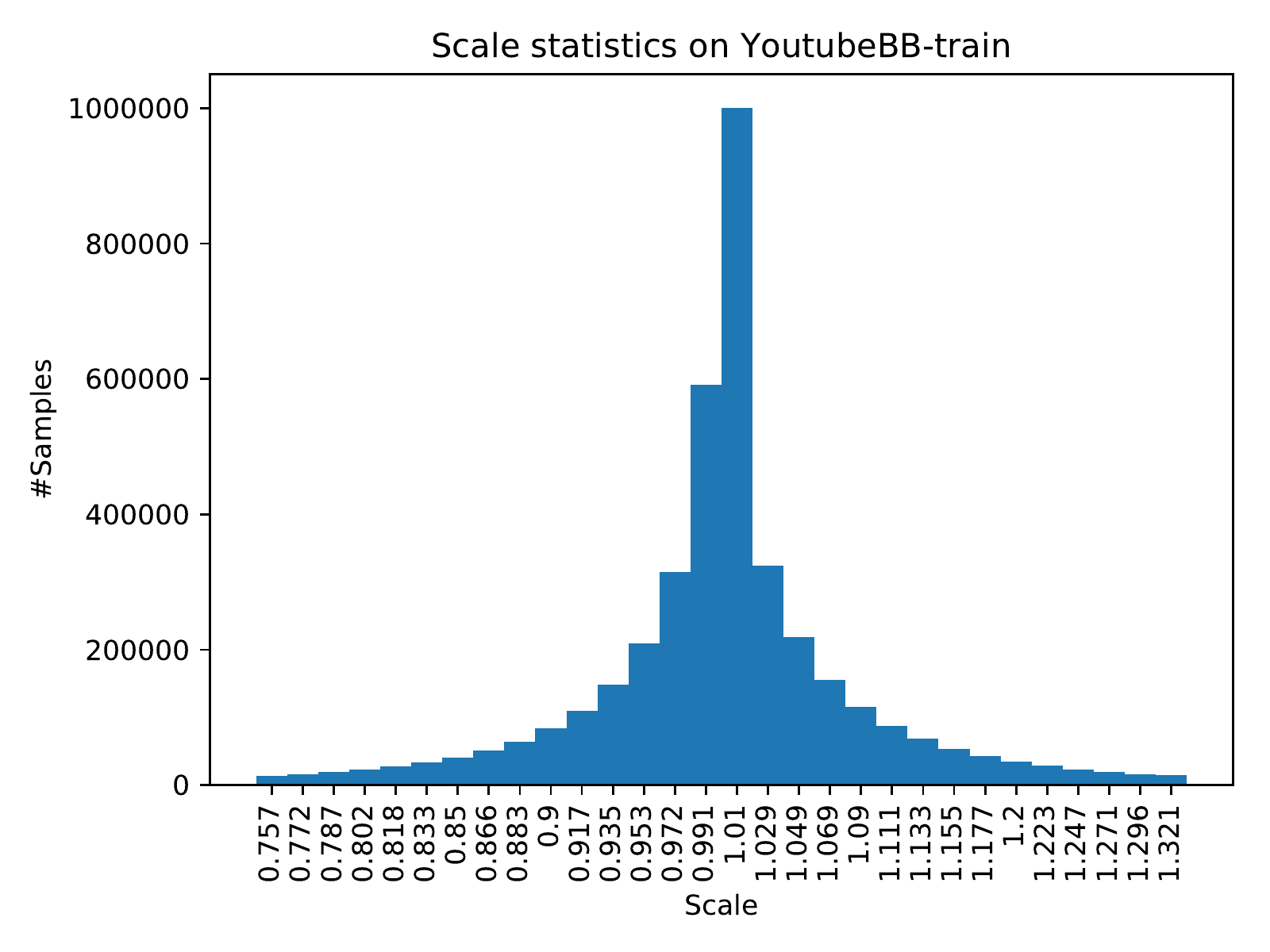}
    &
    \includegraphics[width=\StatW,height=\StatH]{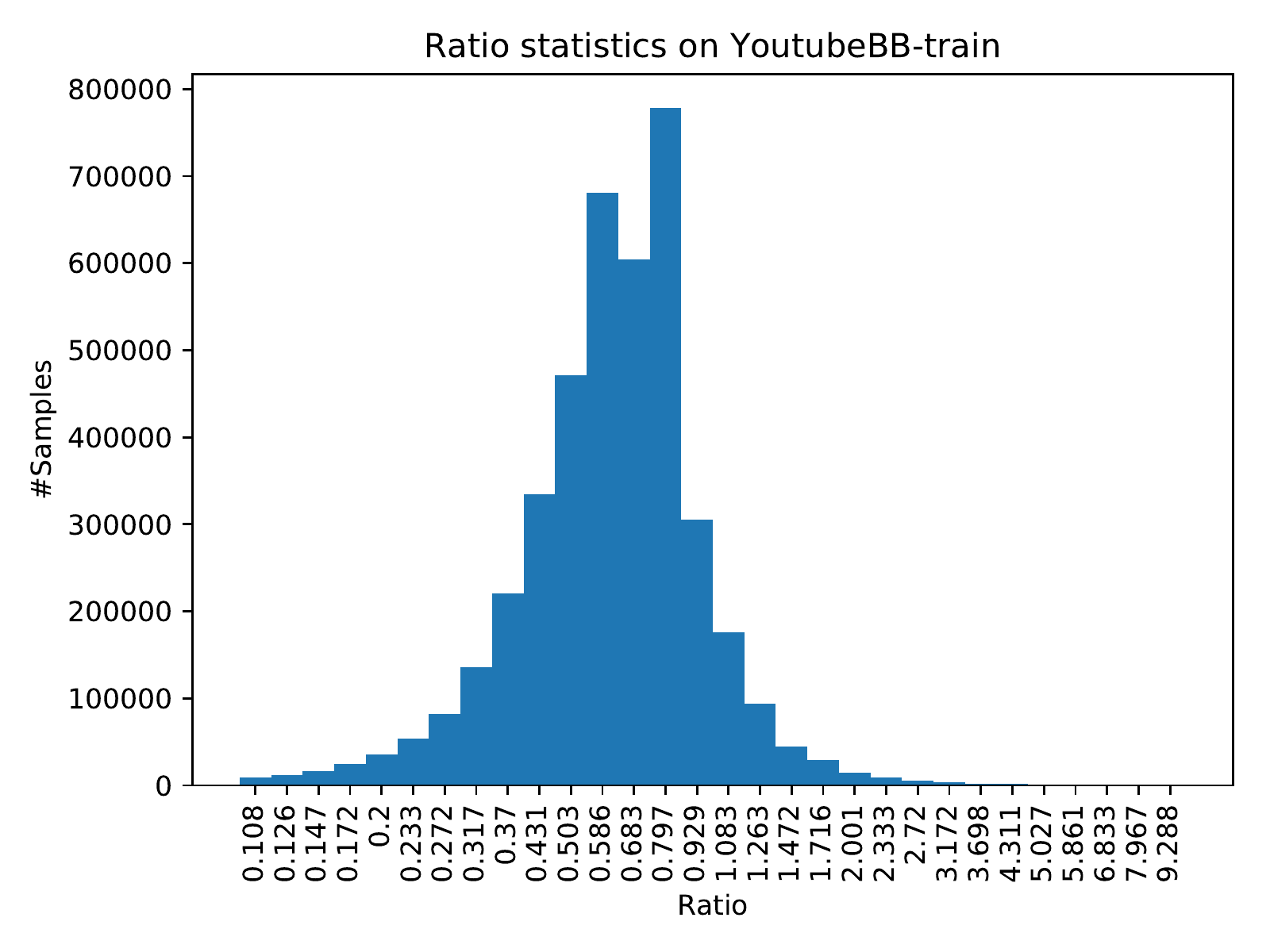}
    &
    \includegraphics[width=\StatW,height=\StatH]{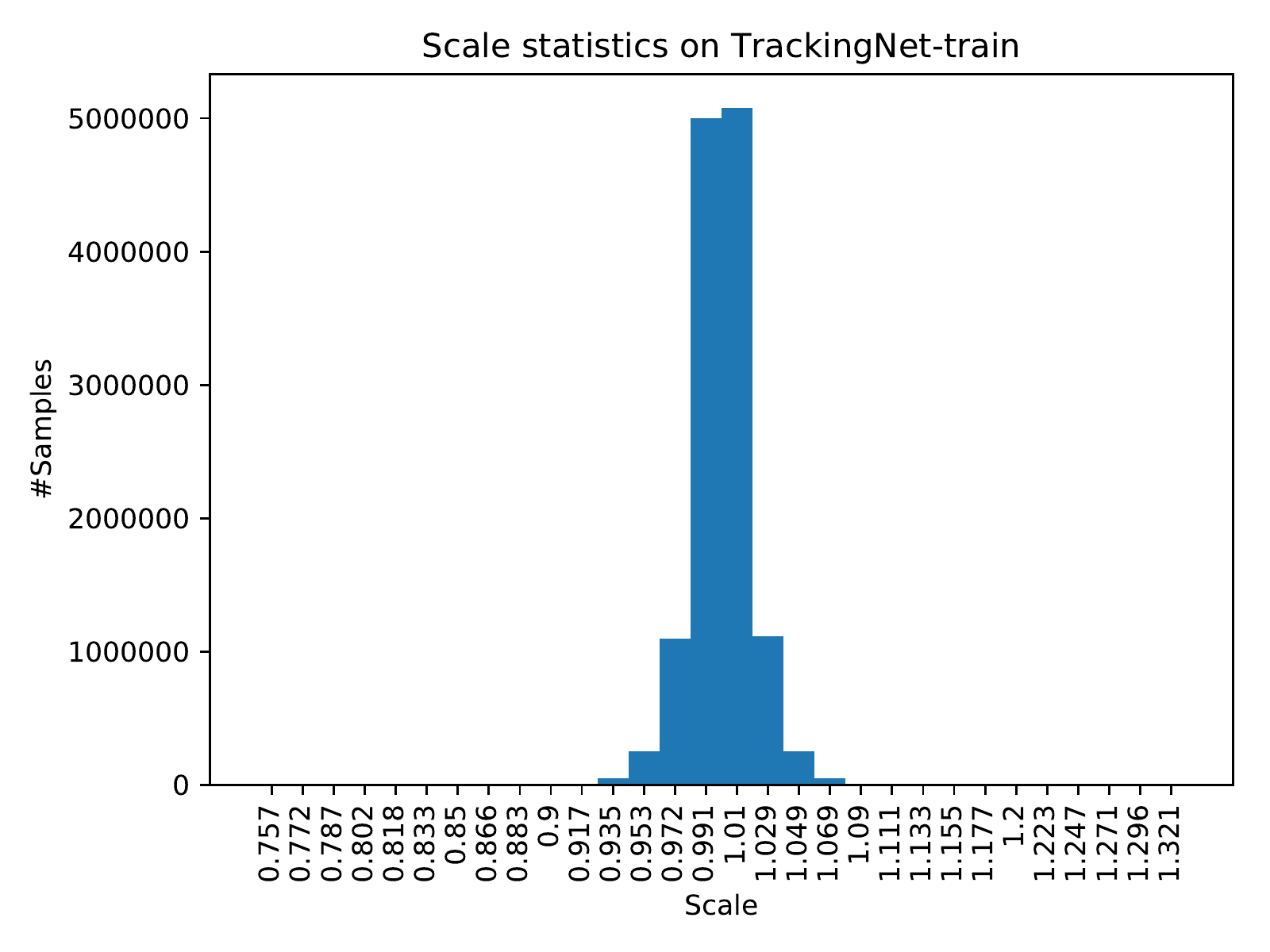}
    &
    \includegraphics[width=\StatW,height=\StatH]{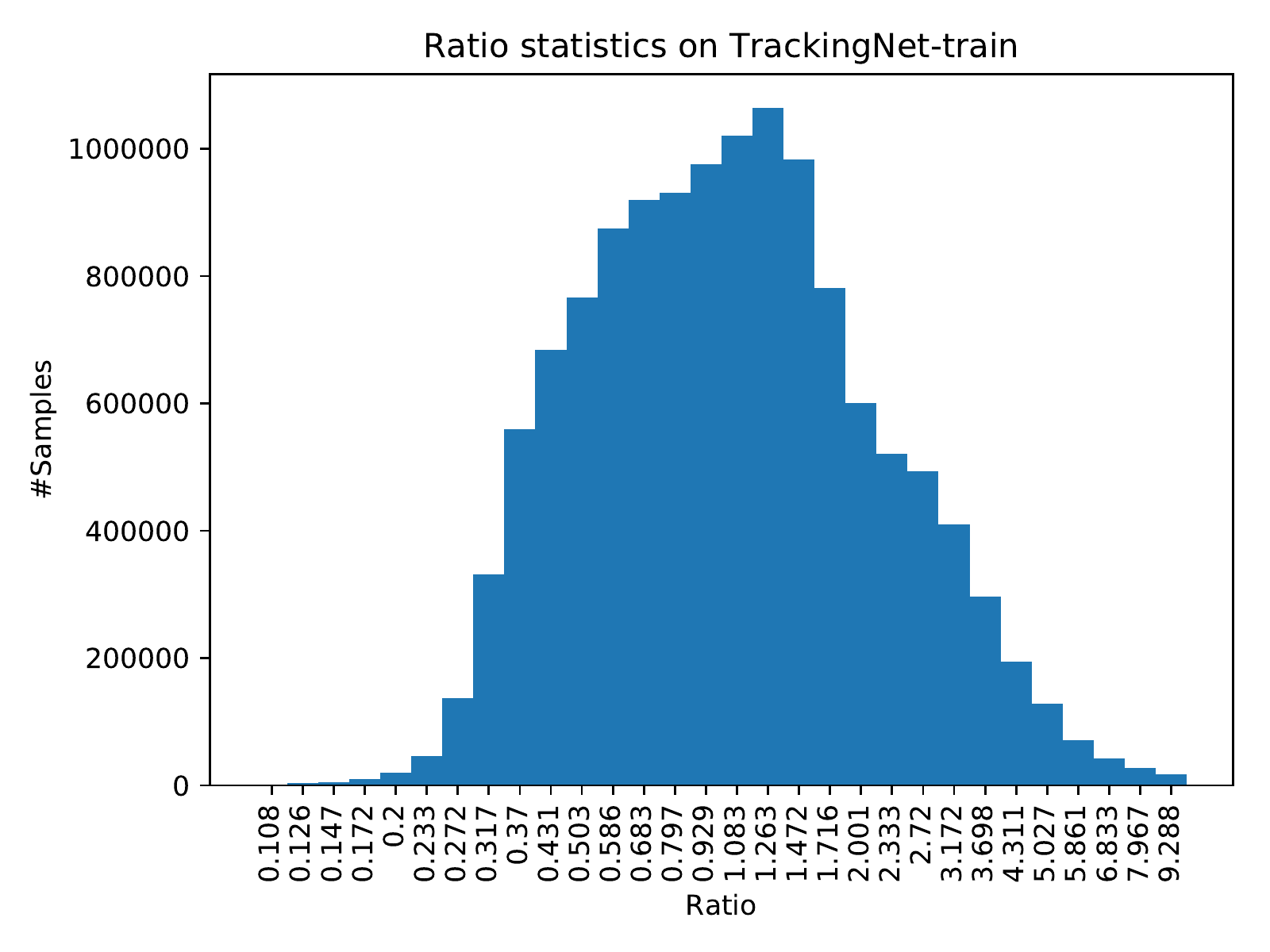}
\\
    \includegraphics[width=\StatW,height=\StatH]{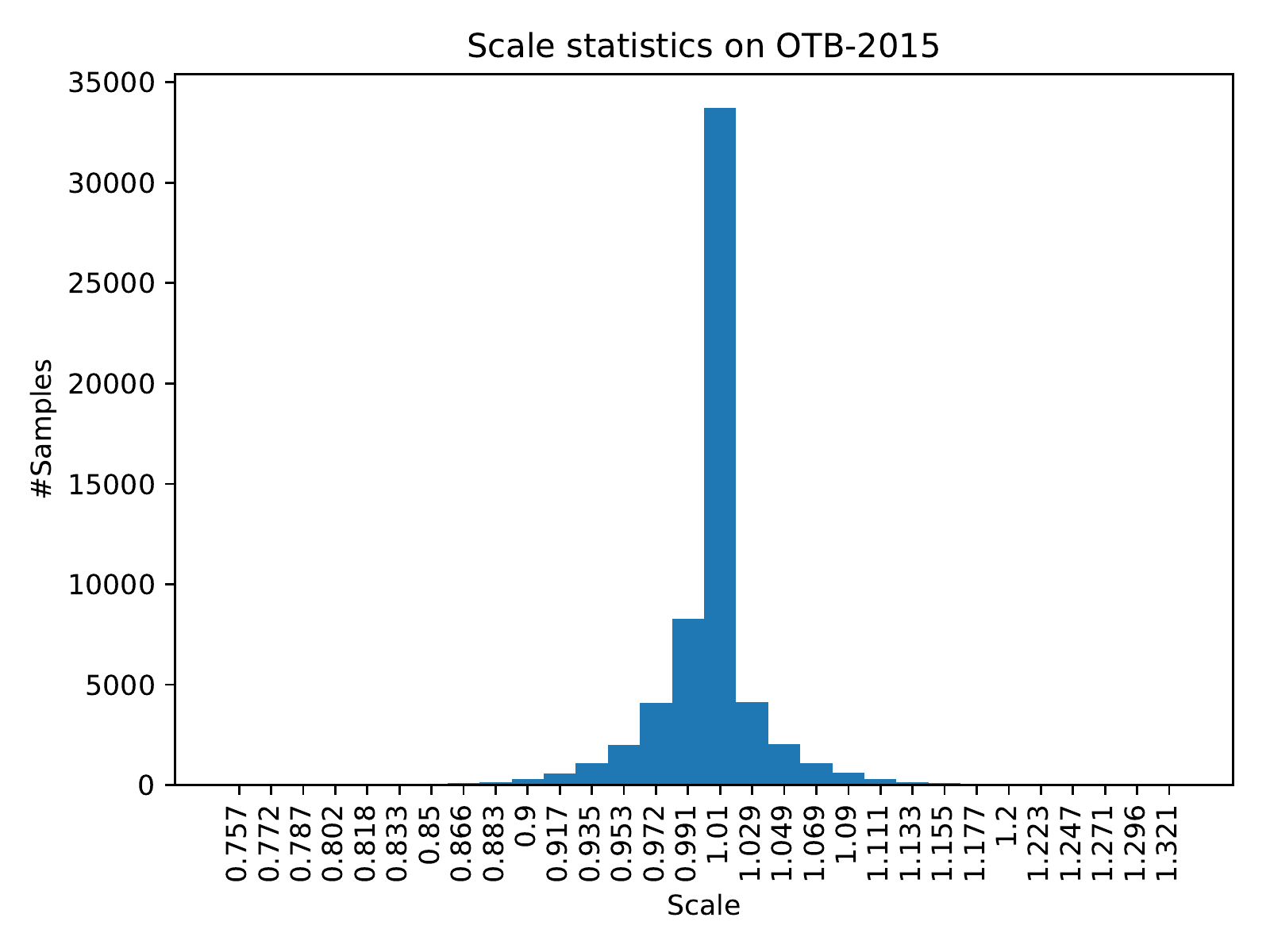}
    &
    \includegraphics[width=\StatW,height=\StatH]{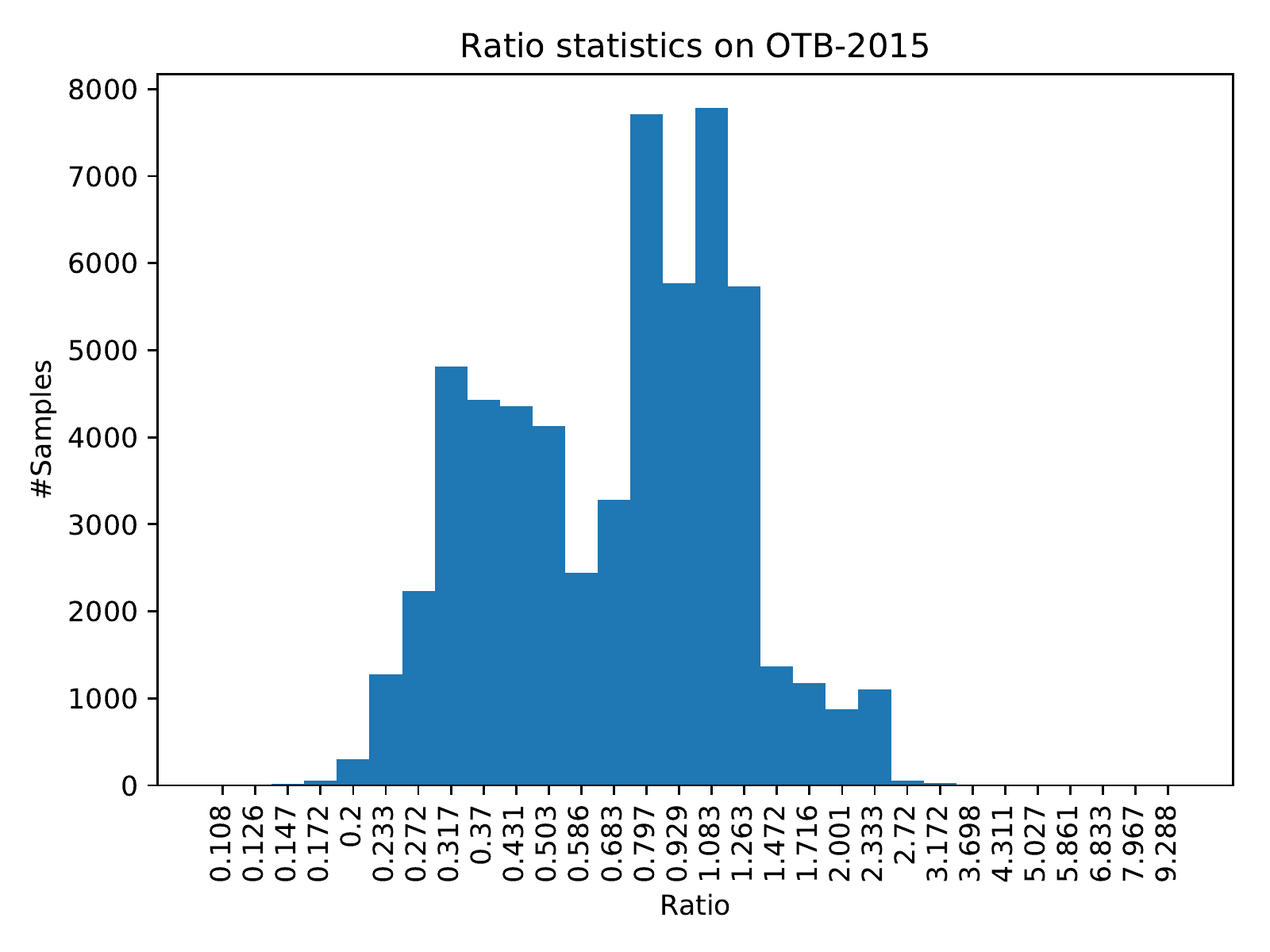}
    &
    \includegraphics[width=\StatW,height=\StatH]{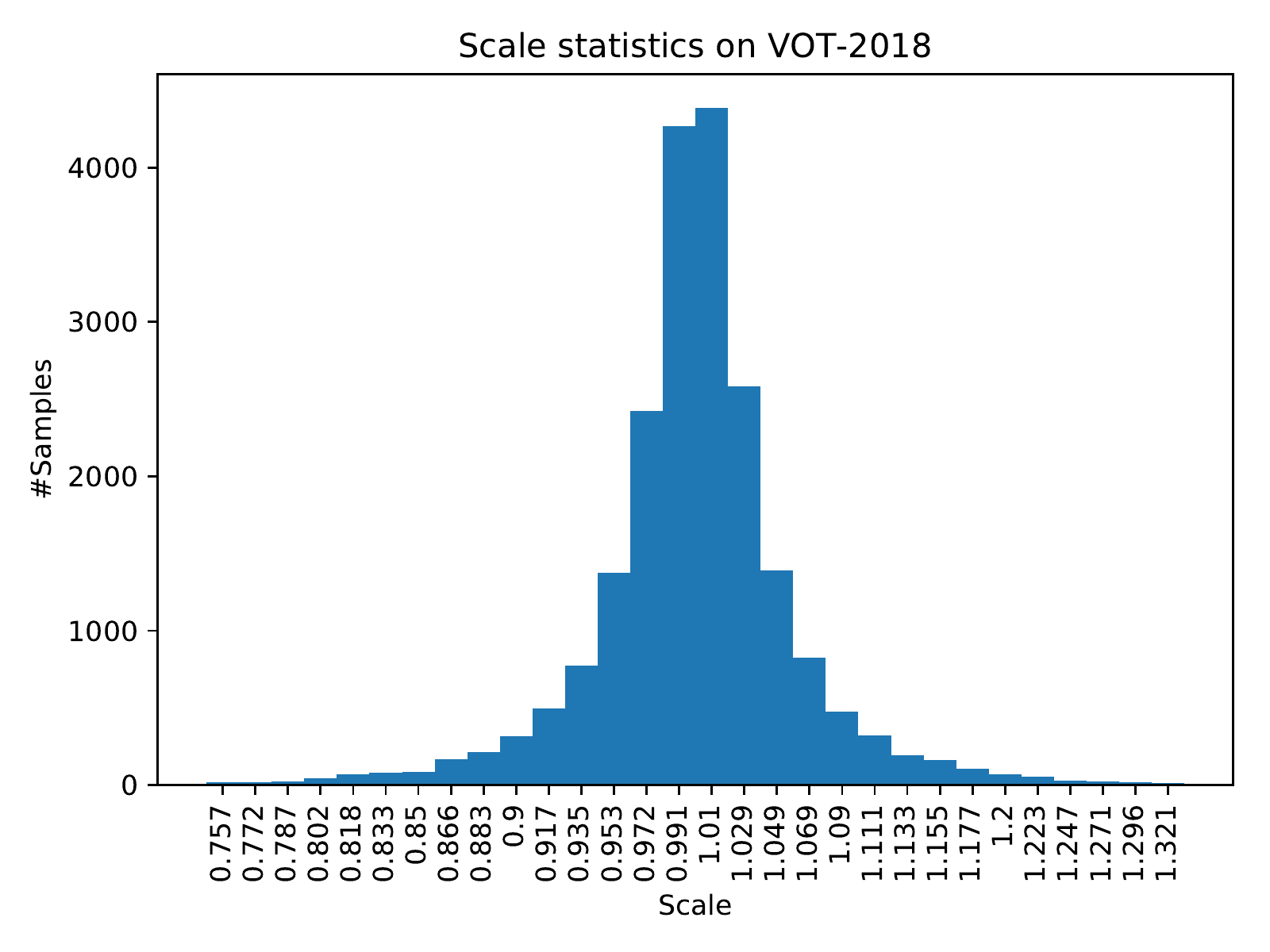}
    &
    \includegraphics[width=\StatW,height=\StatH]{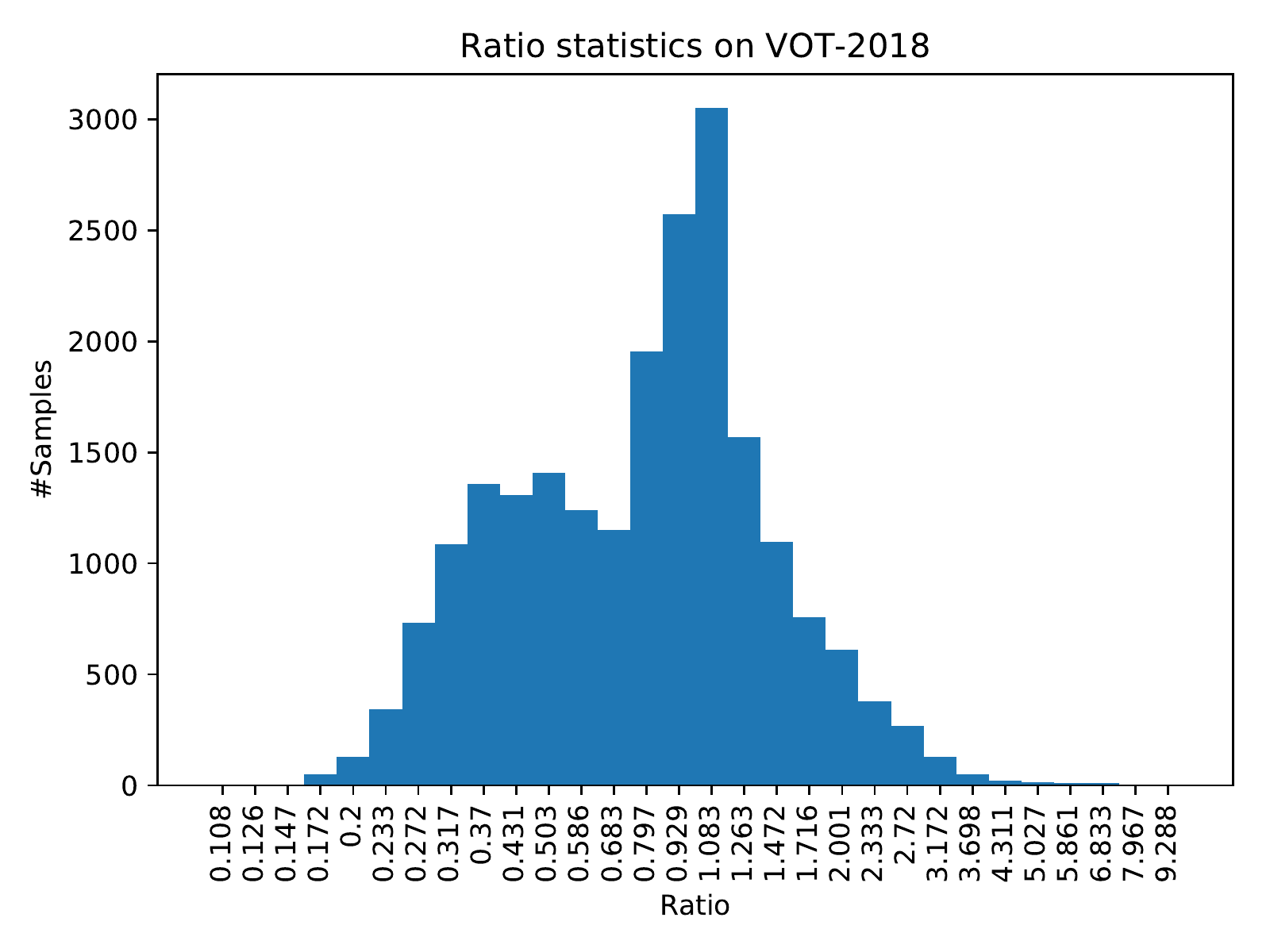}
\end{tabular}
\caption{Scale/ratio statistics on involved datasets. \textit{Scale} refer to the relative scale(i.e. scale change $\tilde{s}_T = s_T / s_{T-1}$ where $s_T = w_T \cdot h_T$) while \textit{ratio} keeps its original definition(weight/height).}
\label{tab:scale/ratio statistics}
\end{table*}

\section{Data Distribution of Different Datasets}
Data distribution varies between different datasets. We collect statistics over scale/ratio distribution across the involved datasets and the results are presented in Table \ref{tab:scale/ratio statistics}. Each dataset does not share the same diversity and the data distribution itself can be a sort of prior knowledge. Seeing that the access to prior knowledge is against the spirit of generic object tracking~\cite{huang2018got}, tracker design should prevent the involvement of factors related to scale/ratio information.

\section{Comment over Anchor-based Scoring: A Maxout Perspective}
Under the post-processing scheme described in Section~\ref{sec:test phase behavior}, the scoring process in anchor-based tracking methods can be viewed as a \textit{maxout}~\cite{10.5555/3042817.3043084} of scores of foreground (scores of anchors) at each pixel location. 

According to the practice in the face detection task~\cite{tang2018pyramidbox}, the maxout of scores of foreground prefers to \textit{recall} and thus avoid false negatives (FNs), while the maxout of scores of background is prone to \textit{select} and thus suppress false positives (FPs). 

In the SOT task, an FP is usually \textit{worse} than an FN. The reason is that FP is likely to cause drift, while an FN just returns low scores at every pixel and the predicted bounding box of the target will be preserved as the same in the last frame due to the cosine window of the aforementioned post-processing (see Section~\ref{sec:test phase behavior} for detail).

In this perspective of maxout, we can see that the avoidance of ambiguity brought by the per-pixel scoring avoids the maxout of scores of foreground, which contributes to the enhancement of the robustness of our proposed SiamFC++ tracker. 
\end{appendices}

\end{document}